\newif\ifOneCol

\ifOneCol
    \documentclass[12pt, draftclsnofoot, onecolumn]{IEEEtran}
\else
    \documentclass[journal,twocolumn]{IEEEtran}
\fi
\usepackage{amsmath,amsfonts}
\usepackage{algorithmic}
\usepackage{algorithm}
\usepackage{array}
\usepackage{textcomp}
\usepackage{stfloats}
\usepackage{url}
\usepackage{verbatim}
\usepackage{graphicx}
\usepackage{cite}
\usepackage{multirow,hhline,amsthm,subfigure,makecell,dsfont,enumitem,color}
\usepackage[T1]{fontenc}

\theoremstyle{definition}
\newtheorem*{definition}{Definition}
\newenvironment{itpars}{\par\itshape}{\par}
\DeclareMathOperator*{\argmax}{argmax}

\begin{document}

\title{Transfer Learning with Reconstruction Loss}

\author{\IEEEauthorblockN{Wei Cui, \IEEEmembership{Student Member,~IEEE}, and Wei Yu, \IEEEmembership{Fellow,~IEEE}} 
\thanks{Manuscript submitted to \emph{IEEE Transactions on Machine Learning in Communications and Networking} on April 6, 2023, revised on October 18, 2023, and accepted on March 24, 2024. The materials in this paper have been presented in part at the IEEE Global Communications Conference (Globecom),
Rio de Janeiro, Brazil, December 2022 \cite{conf}. This work is supported by Natural Sciences and Engineering Research Council (NSERC) of Canada via the Canada Research Chairs Program.  
The authors are with The
Edward S.~Rogers Sr.~Department of Electrical and Computer Engineering,
University of Toronto, Toronto, ON M5S 3G4, Canada 
(e-mails: \{cuiwei2, weiyu\}@ece.utoronto.ca).}}



\maketitle

\begin{abstract}
In most applications of utilizing neural networks for mathematical optimization, a dedicated model is trained for each specific optimization objective. However, in many scenarios, several distinct yet correlated objectives or tasks often need to be optimized on the same set of problem inputs. Instead of independently training a different neural network for each problem separately, it would be more efficient to exploit the correlations between these objectives and to train multiple neural network models with shared model parameters and feature representations. To achieve this, this paper first establishes the concept of \emph{common information}: the shared knowledge required for solving the correlated tasks, then proposes a novel approach for model training by adding into the model an additional reconstruction stage associated with a new \emph{reconstruction loss}. This loss is for reconstructing the common information starting from a selected hidden layer in the model. The proposed approach encourages the learned features to be general and transferable, and therefore can be readily used for efficient transfer learning. For numerical simulations, three applications are studied: transfer learning on classifying MNIST handwritten digits, the device-to-device wireless network power allocation, and the multiple-input-single-output network downlink beamforming and localization. Simulation results suggest that the proposed approach is highly efficient in data and model complexity, is resilient to over-fitting, and has competitive performances. \end{abstract}

\begin{IEEEkeywords}
Transfer learning, feature learning, mathematical optimization, wireless communications, information flow
\end{IEEEkeywords}

\section{Introduction}
Deep learning has gained increasing popularity as a flexible and computationally efficient approach for solving a great variety of mathematical optimization problems, such as resource allocations \cite{hong_spawc,spatial_learn,ensemble_learn,robust_dl}, detection and sensing \cite{unfold,khobahi,tddmimo,sensing}, and so on. In most literature on applying deep learning for solving optimization problems, a specialized neural network is trained from scratch for each individual optimization task. Such an approach requires a large number of training data for obtaining satisfactory performances on each task, and lacks scalability when multiple objectives need to be optimized. However, in many scenarios, there are often multiple optimization problems that are based on the same set of inputs and differ from each other only in terms of their objective functions. In this paper, we exploit the similarities between these optimization tasks and purpose a novel deep learning approach to train neural networks for different tasks in a highly efficient way, in terms of both data and model complexity.

In the machine learning literature, researchers have explored the transfer of knowledge between machine learning models to tackle similar tasks, known as \emph{transfer learning} \cite{survey}. In these works, a model is first fully trained from scratch with abundant training data and computation resources for one task (i.e. the \emph{source task}). When a new task correlated to the source task is presented (i.e. the \emph{target task}) with only a small amount of data available, the trained model is further fine tuned based on the limited available data to solve this new task. Transfer learning is popular in computer vision (CV) \cite{mingsheng, yaroslav, saenko, gopala, srikanth}, natural language processing (NLP) \cite{fuzhen,jundeng,bert,gpt}, and so on.

To better understand transfer learning with neural networks, we interpret the neural network input-to-output computation flow as a two-stage process, i.e. a \emph{feature learning stage} followed by an \emph{optimization stage}:
\begin{enumerate}
\item \emph{Feature learning stage}: the stage where the high-level feature representations are learned. 
\item \emph{Optimization stage}: the stage where the final task-specific outputs are computed based on the learned features. 
\end{enumerate}
Many transfer learning approaches can be viewed as transferring the feature learning stage across neural network models, while each model learns its own task-specific optimization stage. In the mainstream transfer learning research in application fields such as CV and NLP, the inputs in the problems are typically highly structured, while the outputs (or targets) are in much lower dimensions. Consequently, it is clear where the feature learning stage and the optimization stage are within the neural network computation flow. However, conducting transfer learning on general mathematical optimization problems is different. Here, the inputs, the outputs, and their mappings often lack discernible structures, resulting in no clear distinction between the feature learning stage and the optimization stage in the neural network. Consequently, it is difficult to determine where within the trained model the transferable knowledge (in the form of features) is computed, or even if such transferable knowledge exists at all. 

\subsection{Main Contributions}
In this paper, we propose a novel transfer learning approach to explicitly enforce the learning of transferable features within a specific location in the neural network computation flow\footnote{The code for this paper is available at: \newline \url{https://github.com/willtop/Transfer_Learning_with_Reconstruction_Loss}}. Firstly, we establish the following concept for transfer learning: 
\begin{itpars}
    Common Information\footnote{The term ``common information" has also been used in information theory as a similarity measurement between two correlated random variables \cite{commoninfoit}, which is not to be confused with the definition in this paper.}: The information required for specifying and for solving both the source task and the target task.
\end{itpars}
Although common information may be difficult to identify in general, in this paper, we make a key observation that when the source task and the target task share the same input, the problem input itself always forms a (possibly non-strict) super-set of the common information. Therefore, in this paper, we consider transfer learning for problems that share the same input distribution and propose to use the problem input as a general choice for common information. Moreover, this paper also goes beyond using the input as common information and further considers applications where we can extract specific common information with lower dimensional representations. In this case, we can further adopt the proposed approach for these specific representations. 

With the concept of common information established, the proposed transfer learning approach can be described as follows. When training the neural network model on the source task, besides the task-specific loss, we introduce an additional \emph{reconstruction loss} to be minimized jointly: i.e., we let the neural network reconstruct the common information using features from a specific hidden layer (referred to as the \emph{feature layer}) and compute the common information reconstruction loss. Through minimizing this reconstruction loss, we encourage the features learned at the feature layer to be informative about all the correlated tasks that take the same input distribution. To perform transfer learning on a target task, we fix the trained model parameters up to this feature layer as the already-optimized feature learning stage for the model and further train the remaining model parameters on the target task. 

When the proposed approach utilizes a choice of common information that is generic (e.g. the problem inputs), the features learned in the model can be used for all target tasks that have the same input. Essentially, the proposed transfer learning approach is \mbox{\emph{target-task agnostic}}. This is in contrast to some of the prior transfer learning works \cite{shai,fuzhen2,jundeng,meiyu} where the training approach is dedicated to a given source-task and target-task pair. We note that several works also explore the similar idea of encouraging input reconstruction from the model's internal features, e.g., in the field of semi-supervised learning \cite{ranzato,hybridnet}, multi-task learning \cite{guangquan}, or domain adaptation (a sub-category of transfer learning where the input distribution changes from the source to the target domain) \cite{fuzhen2,glorot,dann,pixelda,datl,ghifary}. Nonetheless, these works deal with different problem setups: in semi-supervised learning, the focus is on obtaining quality features and latent representations of the inputs with only limited ground-truth labels; in multi-task learning, the model is trained on multiple task simultaneously using ample training data; in domain adaptation, the task stays the same while the input distribution changes. While the input reconstruction technique in these works all aims to encourage the neural network models to extract salient features, the present paper differs in that we focus on transfer learning and specifically on quick adaptation to new unseen tasks, after the initial training stage, with only limited further adjustments of model parameters. Moreover, this paper goes one step further in expanding the possibilities for the reconstruction target beyond the problem inputs, through introducing the concept of common information.  

For numerical simulations, we first demonstrate the proposed approach through a classical machine learning application: the MNIST handwritten digits classification \cite{mnist}. We formulate the source task and the target task as correlated classification tasks and adopt the fully-connected neural networks. Furthermore, we treat several important and challenging classes of mathematical optimization problems in wireless communications. Although a few works have explored transfer learning on certain wireless communication problems \cite{hyewon,yiyuan,meiyu}, the techniques used are specific to the application settings or objective characteristics. On the other hand, our approach is more general and readily adapts to different problems or objectives. We illustrate this by experimenting over two application scenarios: the power control utility optimization for device-to-device (D2D) wireless networks, as well as the downlink beamforming and localization problems for multiple-input-single-output (MISO) wireless networks. Specifically, for each of the application scenario, we explore transfer learning between a pair of distinct yet correlated objectives: the min rate and sum rate objectives in the D2D networks; and the beamforming gain and the localization accuracy in the MISO networks. Optimization results suggest that the proposed approach achieves better knowledge transfer and mitigates over-fitting on limited target task data more effectively than the conventional transfer learning method.

\subsection{Paper Organization and Notations}
The remaining of the paper is organized as follows. Section~\ref{sec:formulation} formulates the general transfer learning problem and establishes the concept of common information. Section~\ref{sec:prob} introduces the three applications to be studied in details: the MNIST classification problem, the D2D network power control problems, and the downlink MISO network beamforming and localization problems. Section~\ref{sec:method} proposes the novel transfer learning approach for general neural networks, along with the proper selections of the common information for the three applications. The performances of the
proposed method over three applications are presented and analyzed in Section~\ref{sec:result}. 
Finally, conclusions are drawn in Section~\ref{sec:conclusion}.

For mathematical symbols, we use lower-case letters for scalar variables, lower-case bold-face letters for vector variables, and upper-case bold-face letters for matrix variables. By default, we regard any vector as a single-column matrix. We use the superscript $(\cdot)^\text{H}$ to denote the Hermitian transpose of a matrix (or a vector regarded as a single-column matrix). We use $\mathcal{CN}(0,\Sigma)$ to denote the zero-mean circularly-symmetric complex normal distribution, with $\Sigma$ being the covariance matrix. We use $[\cdot]_k$ to denote the \mbox{$k$-th} element of a vector. We use $|\cdot|$ to denote the absolute value of a complex number, and $||\cdot||_2$ to denote the $L^2$ norm of a vector. Lastly, we use the operator $\leftarrow$ to denote the assignment to a specified variable.

\section{Transfer Learning Formulation}\label{sec:formulation}
We first present the general transfer learning formulation studied in this paper. This transfer learning formulation is not restricted to any specific application and is applicable to various application domains and scenarios. Specifically, we introduce the novel concept and the precise definition of \mbox{\emph{common information}}. At the end of this section, we provide a discussion on the fundamental learning objectives and requirements for efficient transfer learning. 

\subsection{General Setup: Source Task and Target Task Optimization}\label{sec:formulation_I}
Among many variants of transfer learning formulations, we focus on the transfer learning setting where the source task and the target task share the same input distribution but differ in their respective objectives. Let $\mathcal{S}$ denote the source task and $\mathcal{T}$ denote the target task. Consider the optimization problems summarized by the following components:
\begin{itemize}
    \item Input parameters $\mathbf{p}$ summarizing all environment information essential for optimization, which follow the same distribution in both $\mathcal{S}$ and $\mathcal{T}$;
    \item Optimization variables for $\mathcal{S}$: $\mathbf{x}_s$;
    \item Objective (or utility) for $\mathcal{S}$: $u_s(\mathbf{x}_s)$;
    \item Optimization variables for $\mathcal{T}$: $\mathbf{x}_t$;
    \item Objective (or utility) for $\mathcal{T}$: $u_t(\mathbf{x}_t)$.
\end{itemize}
If $\mathcal{T}$ and $\mathcal{S}$ are supervised learning tasks, $u_s(\mathbf{x}_s)$ and $u_t(\mathbf{x}_t)$ are also dependent on the ground-truth labels, which we omit in our notations as they are not variables to be optimized.

To optimize for $\mathcal{S}$ and $\mathcal{T}$, we utilize neural networks to compute the optimal values of the optimization variables. Specifically:
\begin{itemize}
    \item Let $\mathcal{F}_{\Theta_s}$ with trainable parameters $\Theta_s$ denote the neural network mapping for optimizing $\mathcal{S}$:
    \begin{align}\label{equ:xs_func}
        \mathcal{F}_{\Theta_s}(\mathbf{p}) = \mathbf{x}_s \:.
    \end{align}
    \item Let $\mathcal{F}_{\Theta_t}$ with trainable parameters $\Theta_t$ denote the neural network mapping for optimizing $\mathcal{T}$:
    \begin{align}\label{equ:xt_func}
        \mathcal{F}_{\Theta_t}(\mathbf{p}) = \mathbf{x}_t \:.
    \end{align}
\end{itemize}

\subsection{Common Information}\label{sec:formulation_II}
Under this transfer learning setting, $\mathcal{S}$ and $\mathcal{T}$ are correlated in the sense that for optimizing $u_s(\mathbf{x_s})$ and $u_t(\mathbf{x_t})$, the information extracted from $\mathbf{p}$ should be similar. To mathematically formalize this concept, we introduce the concept of \emph{common information}, which we denote by $\mathcal{I}(\mathbf{p})$, with the following definition:
\begin{definition}[Common Information]\label{def:commoninfo}
Let the input parameters $\mathbf{p}$ follow the same distribution for the source task $\mathcal{S}$ and the target task $\mathcal{T}$. Let $\mathbf{x}_s^*(\mathbf{p})$ and $\mathbf{x}_t^*(\mathbf{p})$ denote the optimal solution for $\mathcal{S}$ and $\mathcal{T}$ respectively as follows\footnote{Note that by (\ref{equ:xs_func}) and (\ref{equ:xt_func}), both $\mathbf{x}_s$ and $\mathbf{x}_t$ are functions of the input $\mathbf{p}$. Thus, the optimal $\mathbf{x}_s^*$ and $\mathbf{x}_t^*$ are also functions of $\mathbf{p}$.
}:
\begin{align}
\mathbf{x}_s^*(\mathbf{p}) =& \argmax_{\mathbf{x}_s}u_s(\mathbf{x}_s) \\
\mathbf{x}_t^*(\mathbf{p}) =& \argmax_{\mathbf{x}_t}u_t(\mathbf{x}_t) \: .
\end{align}

The common information $\mathcal{I}(\mathbf{p})$ is a function of $\mathbf{p}$ such that it satisfies the following:
\begin{align}
    \exists& \quad\mathcal{F}_1(\cdot),\:\mathcal{F}_2(\cdot) \nonumber\\
    \text{s.t.}& \quad \mathbf{x}_s^*(\mathbf{p}) = \mathcal{F}_1\bigl(\mathcal{I}(\mathbf{p})\bigr) \\
    &\quad \mathbf{x}_t^*(\mathbf{p}) = \mathcal{F}_2\bigl(\mathcal{I}(\mathbf{p})\bigr) \:.
\end{align}
\end{definition}
Essentially, the common information is the information required by both $\mathcal{S}$ and $\mathcal{T}$. We note that for all ($\mathcal{S}$, $\mathcal{T}$) pairs, there exists at least one choice of the common information, which is the problem input $\mathbf{p}$ itself.  

\subsection{Transfer Learning Principles}\label{sec:formulation_III}
Despite the correlations, $\mathcal{S}$ and $\mathcal{T}$ are still two different tasks. For a neural network model exclusively trained on one task, its learned features are not optimal for the other one: some features learned from optimizing $\mathcal{S}$ could be totally irrelevant or even counter-productive for optimizing $\mathcal{T}$. For successful transfer learning between $\mathcal{S}$ and $\mathcal{T}$, the features need to be general, i.e., it should contain more knowledge than required for optimizing just a single task. 

Moreover, due to reasons such as the cost and overhead of data acquisition, the data is assumed to be highly limited for training $\Theta_t$ on $\mathcal{T}$. This assumption is particularly relevant for scenarios where the target task $\mathcal{T}$ is adapted on the fly after the model training and deployment. To effectively learn $\mathcal{F}_{\Theta_t}$ with limited data, we need to utilize the correlation between $\mathcal{S}$ and $\mathcal{T}$ and transfer over the knowledge already learned in $\mathcal{F}_{\Theta_s}$ which is also useful for solving $\mathcal{T}$. In essence, features and representations computed both in $\mathcal{F}_{\Theta_s}$ and in $\mathcal{F}_{\Theta_t}$ should contain the common information $\mathcal{I}(\mathbf{p})$ as defined in Section~\ref{sec:formulation_II}.

\section{Applications of Transfer Learning}\label{sec:prob}
We provide in details three applications under the transfer learning setting described in Section~\ref{sec:formulation_I}. First, in Section~\ref{sec:prob_I}, we modify the canonical machine learning problem, the MNIST handwritten digits classification, to a transfer learning problem. 
The MNIST classification task has long been regarded as a benchmark problem in the machine learning literature. By exploring a transfer learning task for this application, we can reveal the potential of the proposed method for learning rich and complex transferable features. Then, we present two wireless communication applications: the D2D power control problem in Section~\ref{sec:prob_II}; and the MISO beamforming and localization problem in Section~\ref{sec:prob_III}. These two wireless communication problems represent examples of non-convex mathematical optimization problems, which lack specific structures in the inputs and solution mappings and therefore are challenging to solve for conventional transfer learning techniques.

\subsection{MNIST Handwritten Digits Classification}\label{sec:prob_I}
MNIST handwritten digits classification \cite{mnist} is one classical problem explored by many machine learning algorithms. Specifically, the MNIST dataset consists of images of handwritings on single digits from 0 to 9. The original classification task requires predicting the correct digit from each handwriting image. 

We formulate the following transfer learning problem based on this original MNIST digit classification problem. We modify the classification objectives to obtain a source task and a target task. Specifically, let $\mathcal{S}$ be the task of identifying whether the input image represents the digit 1 or not; and let $\mathcal{T}$ be the task of identifying whether the input image represents the digit 8 or not. We select the input images (which are the problem inputs $\mathbf{p}$) to only include handwritings representing 0, 1, or 8, such that for both $\mathcal{S}$ and $\mathcal{T}$, the positive and negative samples are relatively well balanced with a ratio around 1-to-2.

The reason for the specific number selections is as follows: as described in Section~\ref{sec:formulation_III}, $\mathcal{S}$ and $\mathcal{T}$ should be similar and at the meantime with significant differences. Both tasks of identifying the number 1 and identifying the number 8 require the common information of the complete pixel patterns such as edges and corners over the entire input image. Nonetheless, the difference is also apparent, as handwritings of the digit 1 should resemble a linear pattern of pixels throughout the image, while handwrittings of the digit 8 should resemble local circular patterns of pixels. As the defined input set only includes handwrittings of 0, 1, and 8, a machine learning model exclusively trained for solving $\mathcal{S}$ only needs to discover and rely on a simple high-level feature: whether the pixels form a global linear pattern or not. Needless to say, the knowledge in this model would not transfer well to solving $\mathcal{T}$, as the learned feature is not informative for distinguishing handwrittings of 8 from 0. Therefore, an effective transfer learning approach is needed for obtaining transferable features from $\mathcal{S}$ to $\mathcal{T}$.

Corresponding to the notions in Section~\ref{sec:formulation_I}, $\mathbf{x}_s$ and $\mathbf{x}_t$ are the neural network predictions on the probabilities that the handwriting input $\mathbf{p}$ represents the respective digit specified by the corresponding task. Specifically, with $\mathcal{S}$ and $\mathcal{T}$ both being binary classification tasks, we have $\mathbf{x}_s\in[0,1]$ being the probability of $\mathbf{p}$ representing the digit 1; and $\mathbf{x}_t\in[0,1]$ being the probability of $\mathbf{p}$ representing the digit 8. Let $\mathbf{p}_\text{digit}$ be the actual digit the handwriting $\mathbf{p}$ represents. The source task and target task objectives are respectively:
\begin{align}
    u_s(\mathbf{x}_s) =&
    \begin{cases}
    1, &\text{if}\ \mathbf{x}_s\geq 0.5 \ \&\ \mathbf{p}_\text{digit}=1\\
    1, &\text{if}\ \mathbf{x}_s< 0.5 \ \&\ \mathbf{p}_\text{digit}\neq1\\
    0, &\text{otherwise}
    \end{cases} \ ,\label{equ:mnist_us}\\
    u_t(\mathbf{x}_t) =& \begin{cases}
    1, &\text{if}\ \mathbf{x}_t\geq 0.5 \ \&\ \mathbf{p}_\text{digit}=8\\
    1, &\text{if}\ \mathbf{x}_t< 0.5 \ \&\ \mathbf{p}_\text{digit}\neq8\\
    0, &\text{otherwise}
    \end{cases} \ ,\label{equ:mnist_ut}
\end{align}
The definitions in (\ref{equ:mnist_us}) and (\ref{equ:mnist_ut}) represent the classification accuracy, i.e., the average values of $u_s(\mathbf{x}_s)$ and $u_t(\mathbf{x}_t)$ over a set of samples are the percentages of correct predictions across that sample set. 

\subsection{D2D Wireless Network Power Control}\label{sec:prob_II}
For the first wireless communication application, we study the power control problem for D2D wireless networks. Consider a wireless network with $N$ D2D links that transmit independently over the frequency band of bandwidth $w$ with full frequency reuse. Let $\mathbf{G}=\{g_{ij}\}_{i,j\in\{1\dots N\}}$ denote the set of channel gains, with $g_{ij}$ being the channel gain from the \mbox{$j$-th} transmitter to the $i$-th receiver. The power control problem takes the wireless channel state information as the input, so we have:
\begin{align}
    \mathbf{p}\leftarrow\mathbf{G}=\{g_{ij}\}_{i,j\in\{1\dots N\}}\ .
\end{align}
Let $P_i$ denote the maximum transmission power for the $i$-th transmitter. The problem of power control is to find the optimal values of the variables $\mathbf{x}=\{x_i\}_{i\in\{1\dots N\}}$, where $x_i\in[0,1]$ denotes the percentage of maximum power that the $i$-th transmitter transmits at. Under a specific power control solution $\mathbf{x}$, the $i$-th link has the following achievable rate:
\begin{align}
    r_i = w\log\left(1+\frac{g_{ii}P_ix_i}{\sum_{j\neq i}g_{ij}P_jx_j + \sigma^2}\right)\:,
\end{align}
where $\sigma^2$ denotes the background noise power level.

For the transfer learning problem, we consider two link rate utility functions that are important under different application scenarios: 
\begin{itemize}
    \item The sum rate optimization as $\mathcal{S}$:
    \begin{align}\label{equ:sum_rate}
        u_s(\mathbf{x}_s) = \sum_{i=1}^N r_i
    \end{align}
    \item The min rate optimization as $\mathcal{T}$:
    \begin{align}\label{equ:min_rate}
        u_t(\mathbf{x}_t) = \min_{i=1\dots N} r_i
    \end{align}
\end{itemize}
where $\mathbf{x}_s$ and $\mathbf{x}_t$ are the power control solutions for the sum-rate optimization $\mathcal{S}$ and the min-rate optimization $\mathcal{T}$ respectively.

Examining (\ref{equ:sum_rate}) and (\ref{equ:min_rate}), they are correlated in the sense that a set of higher rates over all links leads to a higher objective value. Both objectives would benefit from proper interference mitigation. On the other hand, these two objectives differ significantly in term of fairness among links: (\ref{equ:min_rate}) ensures complete fairness by optimizing the worst link rate; (\ref{equ:sum_rate}) largely ignores fairness since the optimal sum rate might be achieved through heavily utilizing strong links. With this distinction, conducting transfer learning between (\ref{equ:sum_rate}) and (\ref{equ:min_rate}) is challenging, as certain features crucial in optimizing (\ref{equ:sum_rate}) could potentially lead to degraded performance in (\ref{equ:min_rate}).

\subsection{MISO Wireless Network Beamforming and Localization}\label{sec:prob_III}
As a second application in wireless communications, we study the transfer learning from the downlink MISO beamforming task to the localization task. The source task aims to design the optimal downlink beamformers based on the uplink received pilots, while the target task aims to find the locations of the users. These two seemingly unrelated tasks nevertheless share the common input as the estimated channel state information. 

Consider a downlink MISO network of $M$ base stations (BS) collectively serving a single user equipement (UE), where each BS is equipped with $K$ antennas while the UE is equipped with a single antenna. We assume that the locations of all BSs are fixed. We denote the UE location by the coordinate $(x_{\text{UE}}, y_{\text{UE}}, z_{\text{UE}})$, with $x_{\text{UE}}$ and $y_{\text{UE}}$ being unknown and $z_{\text{UE}}$ being fixed, which is usually the case in practical \emph{indoor} scenarios where the UE is located on the ground or at a certain level of known height. We assume reciprocity of uplink and downlink channels, with the set of channel coefficients denoted by $\mathbf{H}=\{\mathbf{h}_m\}_{m\in\{1\dots M\}}$, where $\mathbf{h}_m\in\mathbb{C}^K$ is the vector of channel coefficients from the $K$ antennas of the $m$-th BS to the UE. We use the Rician fading model for modeling the wireless channels, with the channels from the $m$-th BS to the UE modelled as follows:
\begin{align}\label{equ:misochannel}
    \mathbf{h}_m = \rho(d_m)\left(\sqrt{\frac{\epsilon}{1+\epsilon}}\mathbf{h}_m^\text{LOS} + \sqrt{\frac{1}{1+\epsilon}}\mathbf{h}_m^\text{NLOS}\right)
\end{align}
with $d_m$ being the distance from the UE to the $m$-th BS and $\rho(d_m)$ being the associated pathloss as a function over this distance, $\mathbf{h}_m^\text{LOS}\in\mathbb{C}^K$ and $\mathbf{h}_m^\text{NLOS}\in\mathbb{C}^K$ being the channel coefficients for the line-of-sight (LOS) path and non-line-of-sight (NLOS) paths respectively, and $\epsilon$ being the Rician factor as the ratio of power between the LOS and NLOS channel components. Specifically, we model the LOS channel component as:
\begin{align}\label{equ:misochannel_LOS1}
    \mathbf{h}_m^{\text{LOS}} = \mathbf{a}_m(\theta_m, \phi_m) \:,
\end{align}
where $\theta_m$ and $\phi_m$ are the azimuth and elevation angle-of-arrival (AoA) respectively from the UE to the $m$-th BS, and $\mathbf{a}_m\in\mathbb{C}^K$ is the steering vector, which is a function of $\theta_m$ and $\phi_m$. Let $\delta$ and $\lambda$ denote the antenna spacing and the signal wavelength respectively, the $k$-th component of $\mathbf{a}_m$, which corresponds to the $k$-th antenna of the $m$-th BS, is computed as follows:
\ifOneCol
    \begin{align}\label{equ:misochannel_LOS2}
        [\mathbf{a}_m(\theta_m, \phi_m)]_k = e^{\frac{2\pi\delta}{\lambda}\left[i_r(n)\sin(\theta_m)\cos(\phi_m)+i_c(n)\cos(\theta_m)\cos(\phi_m)\right]}\:,
    \end{align}
\else
    \begin{multline}\label{equ:misochannel_LOS2}
        [\mathbf{a}_m(\theta_m, \phi_m)]_k = \\ e^{\frac{2\pi\delta}{\lambda}\left[i_r(n)\sin(\theta_m)\cos(\phi_m)+i_c(n)\cos(\theta_m)\cos(\phi_m)\right]}\:,
    \end{multline}
\fi
with $i_r(n)$ and $i_c(n)$ being the row and column index of the \mbox{$n$-th} antenna in the antenna array respectively. As for the NLOS paths, we use the Rayleigh fading model to model these path components collectively:
\begin{align}\label{equ:misochannel_NLOS}
    \mathbf{h}_m^\text{NLOS}\sim \mathcal{CN}(0, \mathbf{I}) \:.
\end{align}

We assume a maximum transmission power level of $P_{\text{BS}}$ for each BS and $P_{\text{UE}}$ for the UE, and a noise level of $\sigma^2$ both at the BSs and at the UE. We use an uplink pilot stage for the BS to infer the wireless channel, thereby allowing it to perform downlink beamforming or localization tasks. Specifically, the UE sends $T$ uplink pilot signals, which each BS measures through applying a sequence of sensing vectors over its $K$ antennas. Let $\mathbf{c}=\{c^t\}_{t\in\{1\dots T\}}$ be the set of uplink pilot signals sent by UE, and $\mathbf{V}=\{\mathbf{v}_m^t\}_{m\in\{1\dots M\},\ t\in\{1\dots T\}}$ be the set of sensing vectors employed by all the BSs, where $\mathbf{v}_m^t\in\mathbb{C}^K$ is the sensing vector adopted by the $m$-th BS for receiving the $t$-th pilot. According to the power constraint, we have $|c^t|^2=P_{\text{UE}},\forall t$. Also, we require $||\mathbf{v}_m^t||_2^2=1,\forall m,t$. For the $t$-th uplink pilot the UE transmits, the $m$-th BS receives the measurement $r_m^t\in\mathbb{C}$:
\begin{align}\label{equ:pilotmeasures}
    r_m^t = (\mathbf{v}_m^t)^\text{H}\mathbf{h}_mc^t + n_m^t \:,
\end{align}
where $n_m^t\sim\mathcal{CN}(0, \sigma^2\mathbf{I})$ is the noise experienced at the $m$-th BS when receiving the $t$-th uplink pilot transmission through $\mathbf{v}_m^t$. 

We study the transfer learning problem between two tasks: the downlink beamforming task $\mathcal{S}$, and the UE localization task $\mathcal{T}$, both of which are based on the uplink pilot measurements as the input. Specifically, we assume both the UE uplink pilots $\mathbf{c}=\{c^t\}_{t\in\{1\dots T\}}$ and the sensing vectors adopted at each BS $\mathbf{V}=\{\mathbf{v}_m^t\}_{m\in\{1\dots M\},\ t\in\{1\dots T\}}$ are fixed. Correspondingly, we have the problem inputs as: 
\begin{align}
\mathbf{p} \leftarrow \{r_m^t\}_{m\in\{1\dots M\},t\in\{1\dots T\}}\ . 
\end{align}
We now provide detailed descriptions of the source task and the target task as follows.

\subsubsection{Source Task --- Downlink Beamforming}
The task of downlink beamforming focuses on finding the optimal digital downlink beamformers at all BSs to collaboratively maximize the signal power received, or equivalently, the signal-to-noise ratio (SNR), at the UE. Specifically, let $\mathbf{B}=\{\mathbf{b}_m\}_{m\in\{1\dots M\}}$ denote the set of digital beamformers to be optimized, where $\mathbf{b}_m\in\mathbb{C}^K$ is the downlink beamformer employed by the $m$-th BS, with $||\mathbf{b}_m||_2^2=1$. Corresponding to the definition in Section~\ref{sec:formulation_I}, the source-task optimization variables $\mathbf{x}_s$ is the set of beamformers $\mathbf{B}$: 
\begin{align}\label{equ:mimo_xs}
    \mathbf{x}_s^m \leftarrow& \:\mathbf{b}_m \quad\forall m\\
    \mathbf{x}_s\leftarrow&\:\{\mathbf{x}_s^m\}_{m\in\{1\dots M\}}\:.
\end{align}
Given the set of optimized beamformers $\mathbf{B}$, the SNR at the UE for downlink transmission, which is the objective for $\mathcal{S}$, is computed as follows:
\begin{align}\label{equ:beamform_utility}
    u_s(\mathbf{x}_s)=\frac{P_{\text{BS}}\sum_{m=1}^M \mathbf{b}_m^\text{H}\mathbf{h}_m}{\sigma^2}\:.
\end{align}

\subsubsection{Target Task --- Localization}
The task of UE localization focuses on estimating the unknown UE location based on the collection of uplink pilot measurements from all BSs. Specifically, given the problem inputs $\{r_m^t\}_{m\in\{1\dots M\},\ t\in\{1\dots T\}}$, the optimization variables for $\mathcal{T}$, i.e., $\mathbf{x}_t$, are the estimation of the x-coordinate and y-coordinate of the UE location, which we denote as $\hat{x}_{\text{UE}}$ and $\hat{y}_{\text{UE}}$. Therefore, we have: 
\begin{align}\label{equ:mimo_xt}
    \mathbf{x}_t \leftarrow (\hat{x}_{\text{UE}},\:\hat{y}_{\text{UE}})\:.
\end{align}
Naturally, the objective for the localization task is to minimize the location estimation error. Using the Euclidean distance as the metric, the utility function is as follows:
\begin{align}\label{equ:localization_utility}
    u_t(\mathbf{x}_t) = -\sqrt{(x_{\text{UE}}-\hat{x}_{\text{UE}})^2+(y_{\text{UE}}-\hat{y}_{\text{UE}})^2}\:.
\end{align}
Note that we define the utility function as the negative value of the estimation error in Euclidean distance. With this definition, we will aim to maximize the utility function $u_t$, which in turn would minimize the  localization error and lead to a high localization accuracy.

\section{Transfer Learning with Reconstruction Loss}\label{sec:method}
We now present the proposed novel transfer learning method, which is effective for a wide range of problem domains and is applicable to a variety of deep neural network architectures. The proposed method is \emph{target-task agnostic} as long as the target task shares the same problem input. As a result, it can be applied to arbitrary target tasks on the fly with minimal additional training. Furthermore, as compared to the conventional transfer learning approaches, only during the source task training, the proposed method introduces some relatively low amount of additional parameter complexity and computational complexity. During the target task training and testing or model implementations however, the proposed method does not introduce any additional parameters or computational complexity.

\subsection{Information within Neural Network Computation Flow}\label{sec:method_I}
A neural network consists of consecutive hidden layers of neurons computing non-linear functions (i.e. activations), forming a computation flow. For a regular neural network learning one specific input-to-output mapping, the features computed at each hidden layer follow a general \emph{information flow} pattern: from the input to the output, the amount of information describing the input gradually reduces layer by layer, while only the information necessary for predicting the output is maintained \cite{naftali,ravid,saxe}. While being efficient for learning a single mapping, this information flow pattern may not be desirable in the transfer learning setting. Instead, we desire the features learned by the model to be generalizable and retain sufficient information for being transferable to new target tasks. In other words, the learned model features need to contain the common information among tasks of interest in order to be effective for transfer learning.

For transfer learning on CV or NLP applications, it is relatively clear which features are likely to hold the common information. Specifically, with highly structured inputs, the neural network computation flow learned under regular training is likely to have structures already: the entire flow can be divided into a feature learning stage and an optimization stage. Take convolutional neural networks solving CV tasks as examples, the feature learning stage includes the convolution layers that compute general high-level features which contain the common information for many relevant CV tasks (such as edge patterns, pixel intensities, or color gradients over the input image), followed by the optimization stage consisting of fully connected layers that process these high-level features and compute the task-specific outputs (e.g. classification class scores). To conduct transfer learning over correlated CV tasks, the convolutional layers are shared among the models as the feature learning stage, while the fully connected layers of each model are further trained on a per-task basis \cite{yaroslav, gopala, srikanth}.

However, for general mathematical optimization problems, the inputs lack clear structures in most cases. Under regular training methods, the resulting neural network computation flow is not clearly divided by stages, with internal features gradually becoming more and more task-specific layer-by-layer. As the result, it is difficult to identify or explicitly encourage the learning of transferable features, or equivalently, to obtain features that incorporate the desired common information. 

\subsection{Source Task Training with Added Reconstruction Loss}\label{sec:method_II}
To tackle the challenges of transfer learning for mathematical optimization problems, we propose a novel transfer learning approach to encourage the learning of transferable features. Specifically, we first identify a proper selection of the common information for the source task and the target task. When training the neural network model on the source task, on top of the regular task-based loss, we introduce in addition a loss term for the common information reconstruction.

\begin{figure*}[t]
\centering
\includegraphics[width=0.9\textwidth]{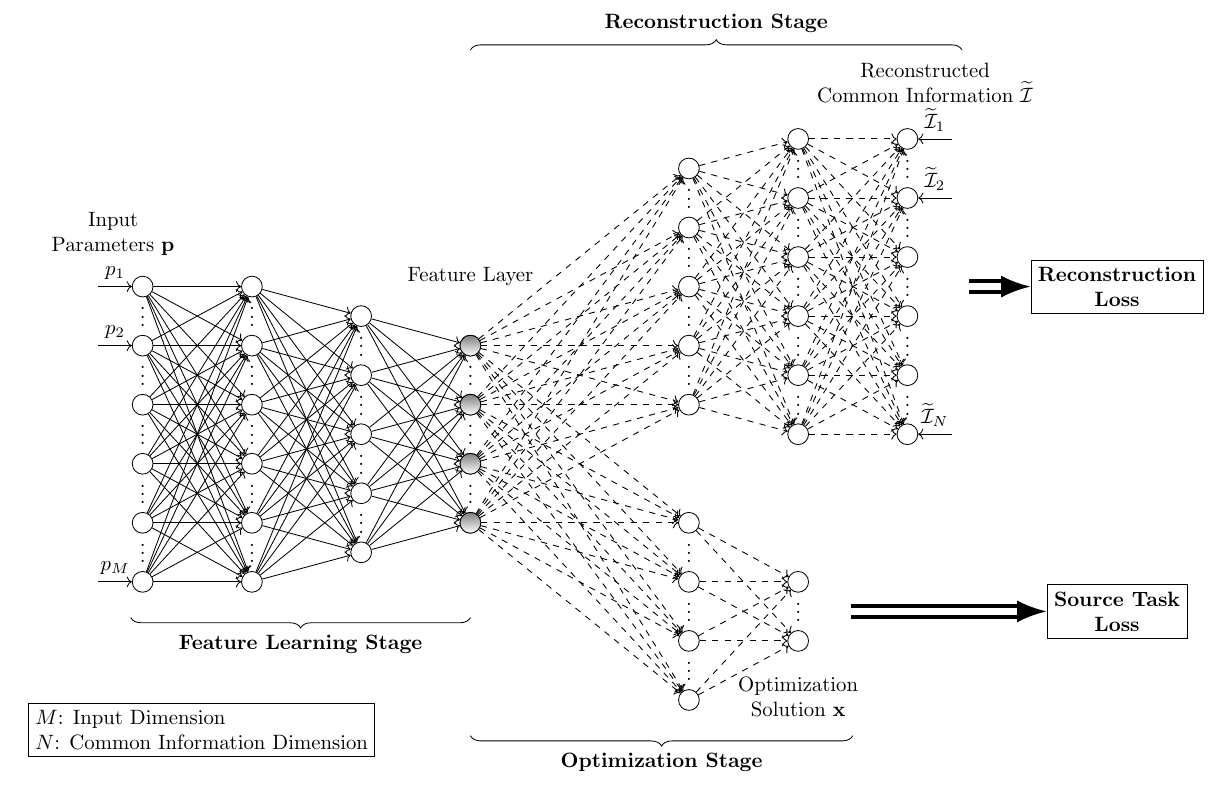}
\caption{Transfer Learning with Reconstruction Loss.}
\label{fig:neuralnet}
\end{figure*}

Fig.~\ref{fig:neuralnet} illustrates the proposed transfer learning approach. We adopt the most general fully-connected neural network architecture. Within the neural network, we select a hidden layer as the \emph{feature layer} where we encourage the transferable features to be computed. Correspondingly, the part of the neural network computation flow from the input layer to the feature layer forms the feature learning stage, while the part of the computation flow from the feature layer to the output layer forms the optimization stage. There is no particular constraint on which hidden layer to select as the feature layer, as long as there is sufficient transformation capacity (by having a sufficient number of hidden layers) both from the input layer to the selected layer and from the selected layer to the output layer. 

From the selected feature layer, we add in a \emph{reconstruction stage} as a separate branch in the computation flow, in parallel to the optimization stage. The reconstruction stage aims to reconstruct the common information $\mathcal{I}(\mathbf{p})$ by using the features in the feature layer. We denote the corresponding reconstruction loss by $\mathcal{L}_R$. Together with the original loss associated with optimizing the source task utility $u_s(\mathbf{x}_s)$, which we denote by $\mathcal{L}_S$, the loss function $\mathcal{L}$ that we use to train $\mathcal{F}_{\Theta_s}$ is:
\begin{align}\label{equ:loss}
    \mathcal{L} = \mathcal{L}_S + \alpha \mathcal{L}_R
\end{align}
where $\alpha$ is the relative weighting scalar between the loss terms. 

With $\mathcal{F}_{\Theta_s}$ trained on $\mathcal{L}$ as in (\ref{equ:loss}), the feature layer computes features that are pertinent to the source task optimization, while also containing knowledge of the common information $\mathcal{I}(\mathbf{p})$. Therefore, these features are highly general and transferable to different tasks that are relevant to the source task. We emphasize that when the choice of $\mathcal{I}(\mathbf{p})$ is generic (e.g., by choosing the problem inputs $\mathbf{p}$ as the common information), the proposed approach is \emph{target-task agnostic}, since no prior knowledge of the target task is needed throughout the training procedure.

\subsection{Transfer Learning by Sharing Feature Layer}\label{sec:method_III}
After training the neural network $\mathcal{F}_{\Theta_s}$ on the source task $\mathcal{S}$ as described in Section~\ref{sec:method_II}, transfer learning on the target task $\mathcal{T}$ is straightforward. We first transfer the subset of the neural network parameters $\Theta_s$ up to the feature layer as the shared feature learning stage to the neural network parameters $\Theta_t$, and leave the remaining parameters in $\Theta_t$ unassigned. When training $\mathcal{F}_{\Theta_t}$, we freeze all the transferred parameters so they would remain unchanged as the already-optimized feature learning stage. With the proposed transfer learning approach, through the transferred feature learning stage, the features computed in the feature layer of $\mathcal{F}_{\Theta_t}$ are already valuable for optimizing $\mathcal{T}$ before starting any target task training. 

For the parameters after the feature layer in $\Theta_t$, we train these parameters with the regular loss associated with the target task utility $u_t(\mathbf{x}_t)$, which we denote as $\mathcal{L_T}$. This further training leads to the task-specific optimization stage in $\Theta_t$. With the number of trainable parameters greatly reduced, along with the fact that the features from the feature layer are also optimized for $\mathcal{T}$, only a small amount of additional training data is needed to obtain a well-performing model $\mathcal{F}_{\Theta_t}$. In essence, the computational complexity of target-task training is significantly reduced when compared to the conventional transfer learning approach.

We emphasize that the selection of $\mathcal{L_T}$ is not required during the source task training phase specified in Section~\ref{sec:method_II}. Instead, we only need to decide the proper $\mathcal{T}$ \emph{after} the target task $\mathcal{T}$ occurs. Therefore, the specification on $\mathcal{L_T}$ does not affect the proposed approach on being target-task agnostic.

For the actual testing or implementations on either $\mathcal{S}$ or $\mathcal{T}$, only the feature learning stage and the corresponding optimization stage of the neural network model need to be executed. Therefore, the proposed approach does not introduce any additional parameters or computational complexity into the model for actual inference. In fact, our approach only increases the model complexity during the source task training. Since in practical scenarios, as the initial stage of the model development, training the model on $\mathcal{S}$ is usually done offline with sufficient time and data resources, the extra complexity from our approach is likely to be negligible.

\subsection{Selecting the Common Information}\label{sec:method_IV}
According to Definition~\ref{def:commoninfo}, selecting the proper common information for a given set of tasks is non-trivial. Fortunately, as mentioned in Section~\ref{sec:formulation_II}, we can always resort to using the problem inputs $\mathbf{p}$ as the common information when no alternative selection is apparent (which also enables the proposed approach to be target task agnostic, as previously discussed). In the following, we examine the three applications for numerical simulations and propose a proper choice for the common information for each of the applications.

\subsubsection{MNIST Digits Classification}
As described in Section~\ref{sec:prob_I}, $\mathcal{S}$ focuses on identifying the digit 1 while $\mathcal{T}$ focuses on identifying the digit 8, over images of handwritings for three digits: 0, 1, and 8. With the inputs $\mathbf{p}$ being images as high-dimensional data, the common information among tasks of identifying different digits is highly complex and involves detecting various pixel patterns and their relative locations in the images. Summarizing the common information in a concise form is difficult if not impossible. Therefore, we select the original problem inputs $\mathbf{p}$, i.e. the images of handwritings, as the common information for this application:
\begin{align}
    \mathcal{I}(\mathbf{p})\leftarrow\mathbf{p}
\end{align}

\subsubsection{D2D Wireless Network Power Control}
With $\mathcal{S}$ focusing on the sum-rate maximization and $\mathcal{T}$ focusing on the min-rate maximization, both power control tasks largely depend on the mutual interference among all the $N$ D2D links. Therefore, the entire set of $N^2$ wireless channels (with $N$ direct-link channels and $N\times(N-1)$ cross-link channels) need to be considered when optimizing both $\mathcal{S}$ and $\mathcal{T}$. Correspondingly, we select all the channel gains $\mathbf{G}=\{g_{ij}\}_{i,j\in\{1\dots N\}}$, which are the problem inputs $\mathbf{p}$, as the common information for this application:
\begin{align}
    \mathcal{I}(\mathbf{p})\leftarrow\mathbf{p}
\end{align}

\subsubsection{MISO Wireless Network Beamforming and Localization}
Unlike the above-mentioned two applications, the common information between $\mathcal{S}$ and $\mathcal{T}$ for this application can be summarized in a concise form. Both downlink beamforming and UE localization implicitly require channel estimation. With the channel model as in (\ref{equ:misochannel}), apart from the unknown stochastic NLOS paths as in (\ref{equ:misochannel_NLOS}), the downlink channels are largely determined by the following geometric parameters:
\begin{enumerate}[label=(\roman*)]
    \item The set of distances from the UE to all $M$ BSs $\{d_m\}_{m\in\{1\dots M\}}$.\label{factor1}
    \item The set of azimuth AoAs from the UE to all $M$ BSs $\{\theta_m\}_{m\in\{1\dots M\}}$.\label{factor2}
    \item The set of elevation AoAs from the UE to al $M$ BSs $\{\phi_m\}_{m\in\{1\dots M\}}$.\label{factor3}
\end{enumerate}
For the source task $\mathcal{S}$ on downlink beamforming, the optimal beamformers that maximize (\ref{equ:beamform_utility}) are the beamformers designed to be perfectly aligned with the downlink channels. Since the geometric parameters \ref{factor1}-\ref{factor3} completely determine the deterministic components within the channels as in (\ref{equ:misochannel})-(\ref{equ:misochannel_LOS2}), the optimization for $\mathbf{x}_s$ is also largely dependent on the knowledge of these geometric parameters. On the other hand, for the target task $\mathcal{T}$ on the UE localization, the UE location estimation ($\hat{x}_{\text{UE}}$, $\hat{y}_{\text{UE}}$, $\hat{z}_{\text{UE}}$) can be obtained through the technique of triangulation from the fixed locations of the BSs, also by using the geometric parameters \ref{factor1}-\ref{factor3}. Therefore, these geometric parameters collectively can serve as a good choice for the common information:
\begin{align}\label{equ:miso_ci}
    \mathcal{I}(\mathbf{p})\leftarrow\{d_m, \theta_m, \phi_m\}_{m\in\{1\dots M\}}\:.
\end{align}

As explained previously, the common information is only required at the source task training stage: it is used as the targets for the reconstruction loss $\mathcal{L_R}$ in (\ref{equ:loss}), as shown in Fig.~\ref{fig:neuralnet}. For the target task training, and more importantly, the actual testing of the model, there is no need for collecting the common information. Correspondingly, for the MISO application, we only need to prepare the geometric parameters \ref{factor1}-\ref{factor3} in the source task training set.

\section{Numerical Simulations}\label{sec:result}
We present the numerical simulation results of the proposed transfer learning method on each of the three applications introduced in Section~\ref{sec:prob}. For each application, we introduce two neural network based benchmarks. To illustrate the effectiveness of the proposed method through comparisons, the neural network models trained under each of the two benchmark methods share the identical network architecture and specification of hidden neurons for the feature learning stage and the optimization stage as the model trained with the proposed approach. The two neural network based benchmark methods are as follows:
\begin{itemize}
    \item \emph{Conventional Transfer Learning}: Train $\mathcal{F}_{\Theta_s}$ for the source task on the loss $\mathcal{L_S}$, then transfer all the parameters in $\Theta_s$ up to the feature layer to $\Theta_t$, followed by training the parameters after the feature layer in $\Theta_t$ on $\mathcal{L_T}$.  
    \item \emph{Regular Learning}: Train $\mathcal{F}_{\Theta_s}$ and $\mathcal{F}_{\Theta_t}$ on the loss $\mathcal{L_S}$ and $\mathcal{L_T}$ respectively, without any knowledge transfer (i.e. parameters sharing) in between.
\end{itemize}
In the following, we refer to the proposed transfer learning method as \emph{Transfer Learning with Reconstruction}. Note that as discussed earlier in this paper, the first benchmark method listed above, which we refer to as the conventional transfer learning method, is the most popular transfer learning method adopted in the literature \cite{yaroslav, gopala, srikanth, fuzhen,jundeng,bert,gpt}.

We emphasize that by definition, both the conventional transfer learning method and the regular learning method lead to the same training updates on $\Theta_s$ in the source task training (since both methods update the entire set of $\Theta_s$ through gradient descents solely on $\mathcal{L_S}$). Thus, the model performances on the source task utility $u_s$ would be the same for these two methods. Correspondingly, for the simulation results below, we present just one value of $u_s$ as the performance achieved by both benchmark methods.

One notable advantage of applying transfer learning, and especially the proposed transfer learning approach, is to mitigate over-fitting during training on limited target task data. On the other hand, early stopping \cite{earlystop} based on validation loss is also a popular technique for combating over-fitting, which however requires a sufficiently large validation set that is often not available for the target task. Nonetheless, for our simulations, we reserve large validation sets on $\mathcal{T}$ and employ early stopping when training each competing neural network model. Correspondingly, each model is evaluated at its best possible performance for each training method, and thus the performance margins fully demonstrate the effectiveness of the knowledge transfer by the proposed method.

\subsection{Transfer Learning MNIST Digits Classification}\label{sec:result_I}
\subsubsection{Simulation Settings}\label{sec:result_I_I}
We take the original MNIST training and evaluation data sets, and keep only the images of handwritings on the digit 0, 1, or 8, as discussed in Section~\ref{sec:prob_I}. To better understand the target task training data efficiency from each method (which shows how effective the transfer learning is), we conduct training and testing under two data-set specifications: a data set specification with highly limited target task training data, referred to as \emph{Data-Spec A}, and a data set specification with significantly more target task training data (around 10-times the size of that in Data-Spec A), referred to as \emph{Data-Spec B}. For Data-Spec A, we divide all the training data according to a 95\%-5\% split for data used in $\mathcal{S}$ and $\mathcal{T}$ respectively. Furthermore, we adopt a 70\%-30\% training-validation split for data in $\mathcal{S}$, and a 10\%-90\% training-validation split for data in $\mathcal{T}$. For Data-Spec B, we divide all the training data according to a 90\%-10\% split for data used in $\mathcal{S}$ and $\mathcal{T}$ respectively. Furthermore, we adopt a 70\%-30\% training-validation split for data in $\mathcal{S}$, and a 50\%-50\% training-validation split for data in $\mathcal{T}$. These data split ratios used in two data set specifications result in the data set sizes shown in Table~\ref{tab:mnist_dataset}. The unusual training-validation split ratios on $\mathcal{T}$ in both Data-Spec A and Data-Spec B are chosen for two reasons: to cater to the assumption that the training data is usually limited in the target task and to ensure a sufficiently large validation set for accurate early stopping in the target task training.

\begin{table}[t]
\caption{MNIST Data Set Specifications}
\begin{center}
\begin{tabular}{|c|c|c|c|}
\hline
\makecell{\topstrut \textbf{Data Set}\\ \textbf{Specification}} & \textbf{Task} & \makecell{\topstrut \textbf{Training Set}\\ \textbf{Samples}} & 
\makecell{\topstrut \textbf{Validation Set}\\ \textbf{Samples}}\\
\hline
\multirow{2}*{Data Spec A} & \topstrut $\mathcal{S}$
&  12313 & 5277 \\ 
\cline{2-4}
& \topstrut $\mathcal{T}$ 
&  92 & 834* \\ 
\hhline{|=|=|=|=|}
\multirow{2}*{Data Spec B} & \topstrut $\mathcal{S}$
&  11664 & 5000 \\ 
\cline{2-4}
& \topstrut $\mathcal{T}$ 
&  926 & 926* \\ 
\hline
\end{tabular}
\label{tab:mnist_dataset}
\end{center}
{\footnotesize $^*$
A large validation set is used here to ensure accurate early stopping.}
\end{table}

As already mentioned in Section~\ref{sec:method_I}, with image inputs, convolutional neural networks naturally form computation flows that have clear structural distinction between the feature learning stage and the optimization stage. However, for more general mathematical optimization problems, there is no established neural network architecture that learns and provides computation flows with such clear distinctions, which is the challenge that the proposed method aims to address. As this MNIST digit classification problem is used for illustrative purpose, the image inputs of the problem are treated as general inputs without clear spatial structures to be exploited directly. Specifically, we flatten the image inputs into one-dimensional vectors\footnote{We down-sample each image to 10$\times$10 pixels before flattening them to vectors for maintaining manageable dimensions. Simulation results suggest that such down-sampling only mildly affect the classification performances from all methods.} and adopt the fully-connected neural network architecture for all neural network models. 

The same overall neural network specification is used by all the competing methods, as shown in Table~\ref{tab:mnist_neuralnet}. We use the same specification for the feature learning stage in $\Theta_t$ and $\Theta_s$ to support the transfer of trained parameters. Furthermore, in this application, since $\mathbf{x}_s$ and $\mathbf{x}_t$ have the same dimension (i.e. a single scalar output), we also use the same specification for the optimization stage in $\Theta_s$ and $\Theta_t$. We note that 25 features are used in the feature layer for each neural network model. Furthermore, each neural network outputs a single value between 0 and 1 (enforced by the sigmoid non-linearity), as the probability of the input handwriting image representing the digit 1 for $\mathcal{T}$ or the digit 8 for $\mathcal{S}$. 

In terms of the selections of the source-task loss function $\mathcal{L_S}$ and the target-task loss function $\mathcal{L_T}$, directly optimizing $u_s$ or $u_t$ as in (\ref{equ:mnist_us}) or (\ref{equ:mnist_ut}) is not feasible, since (\ref{equ:mnist_us}) (or  (\ref{equ:mnist_ut})) is not a differentiable function over $\mathbf{x}_s$ (or $\mathbf{x}_t$), and therefore no gradient can be derived for neural network model parameter updates. Instead, we use the popular cross-entropy function as the task-based loss functions. With $\mathbf{p}_\text{digit}$ being the true digit for $\mathbf{p}$, the (supervised-learning) loss functions for both tasks are as follows:
\ifOneCol
    \begin{align}
        \mathcal{L_S} =& -\mathds{1}(\mathbf{p}_\text{digit}=1)\log(\mathbf{x}_s)-\mathds{1}(\mathbf{p}_\text{digit}\neq1)\log(1-\mathbf{x}_s) \:, \\
        \mathcal{L_T} =& -\mathds{1}(\mathbf{p}_\text{digit}=8)\log(\mathbf{x}_t)-\mathds{1}(\mathbf{p}_\text{digit}\neq8)\log(1-\mathbf{x}_t) \:,
    \end{align}
\else
    \begin{multline}
        \mathcal{L_S} = -\mathds{1}(\mathbf{p}_\text{digit}=1)\log(\mathbf{x}_s)\\-\mathds{1}(\mathbf{p}_\text{digit}\neq1)\log(1-\mathbf{x}_s) \:,
    \end{multline}
    \begin{multline}
        \mathcal{L_T} = -\mathds{1}(\mathbf{p}_\text{digit}=8)\log(\mathbf{x}_t)\\-\mathds{1}(\mathbf{p}_\text{digit}\neq8)\log(1-\mathbf{x}_t) \:,
    \end{multline}
\fi
where $\mathds{1}(\cdot)$ denotes the standard binary indicator function. We use $\alpha=5$ in (\ref{equ:loss}) for the loss $\mathcal{L}$, which provides the best performance for the proposed approach in the process of hyper-parameter tuning.

\begin{table}[t]
\caption{MNIST Neural Network Architecture}
\begin{center}
\begin{tabular}{|l|c|c|}
\hline
\topstrut\textbf{Stages} & \textbf{Layers} & \textbf{Number of Neurons}\\
\hline
\multirow{2}*{\shortstack[l]{Feature Learning\\ Stage ($\Theta_s$ or $\Theta_t$)}}
& \topstrut 1st & 50 \\ \cline{2-3}
& \topstrut Feature Layer & 25 \\ 
\hline
\multirow{2}*{\shortstack[l]{Optimization \\ Stage ($\Theta_s$ or $\Theta_t$)}} 
& \topstrut 1st & 10 \\ \cline{2-3}
& \topstrut Output Layer & 1 \\
\hline
\multirow{2}*{\shortstack[l]{Reconstruction \\ Stage (only $\Theta_s$)}} 
& \topstrut 1st & 50 \\ \cline{2-3}
& \topstrut Reconstruct Layer & 100* \\ 
\hline
\end{tabular}
\label{tab:mnist_neuralnet}
\end{center}
{\footnotesize $^*$
The output dimension corresponds to the down-sampled image dimension.}
\end{table}

\subsubsection{Transfer Learning Performances}
We present the results of the classification accuracy by all the competing methods, trained and evaluated under both data-set specifications, in Table~\ref{tab:mnist_results}. First, the performances of all the methods are about the same on $\mathcal{S}$ (under both data-set specifications), indicating that with the proposed method, the source task performance trade-off due to optimizing the extra reconstruction loss $\mathcal{L}_R$ is minimal. At this very small cost, the performance advantages that our approach achieves on the target task are significant as evident by the noticeable margins obtained on the classification accuracies on $\mathcal{T}$. 

We next examine the evaluation performances on $\mathcal{T}$ under Data-Spec A: the proposed approach achieves the best performance at a 97.4\% prediction accuracy, with a $4.4\%$ and $2.4\%$ performance margins over two benchmark methods. To understand the significance of these margins, we further compare the evaluation results on $\mathcal{T}$ across two data-set specifications. As shown in Table~\ref{tab:mnist_results}, the performance of the proposed approach under Data-Spec A matches that of the two benchmark methods under Data-Spec B. Essentially, with the number of training samples in $\mathcal{T}$ being 92 and 926 under Data-Spec A and B respectively, the results show that to reach the same prediction accuracy that our approach achieves with less than 100 samples, a 10-fold increase of training data on the target task is needed for the two benchmark methods. This comparison result validates the significance of the reported performance margins under Data Spec A, and suggests that our approach indeed achieves high data efficiency in training as a result from effective knowledge transfer from $\mathcal{S}$ to $\mathcal{T}$. Lastly, focusing on the classification accuracies on $\mathcal{T}$ under Data-Spec B, with increased available training data, the conventional transfer learning approach has already lost its edge against the regular learning method, while our proposed approach still produces the best result. Overall, these results indicate that the proposed approach indeed effectively addresses the challenge of transfer learning when the neural network lacks structures in its information flow, through explicitly enforcing the learning of transferable features.  

\begin{table*}[t]
\caption{MNIST Transfer Learning Performances}
\begin{center}
\begin{tabular}{|c|c||c|c|}
\hline
\topstrut \textbf{Task} & \textbf{Method} & \textbf{Accuracy under Data Spec A} & \textbf{Accuracy under Data Spec B}\\
\hline
\multirow{3}*{$\mathcal{S}$: Digit 1}
& \topstrut Regular Learning & \multirow{2}*{99.7\%} & \multirow{2}*{99.7\%} \\\cline{2-2}
& \topstrut Conventional Transfer Learning & & \\\cline{2-4}
& \topstrut Transfer Learning with Reconstruction & 99.7\% & 99.6\% \\
\hhline{|=|=||=|=|}
\multirow{3}*{$\mathcal{T}$: Digit 8}
& \topstrut Regular Learning & 93.0\% & 97.8\% \\\cline{2-4}
& \topstrut Conventional Transfer Learning & 95.0\% & 97.1\% \\\cline{2-4}
& \topstrut Transfer Learning with Reconstruction & 97.4\% & 98.6\% \\
\hline
\end{tabular}
\label{tab:mnist_results}
\end{center}
\end{table*}

\subsection{Transfer Learning on D2D Wireless Network Power Control}\label{sec:result_II}
\subsubsection{Simulation Settings}\label{sec:result_II_I}
We simulate each wireless network containing $N=10$ links randomly deployed in a 150m$\times$150m region. We first generate the locations of the transmitters uniformly within the region, and then generate the locations of the receivers such that the direct-channel transceiver distances follow a uniform distribution in the interval of 5m$\sim$25m. We impose a minimum of 5m distance between any interferring transmitter and receiver. We assume each transmitter has the maximum transmission power of 30dBm with a direct-channel antenna gain of 6dB, while the noise level is -150dBm/Hz. We assume an available bandwidth of 5MHz with full frequency reuse across the entire wireless network.

To simulate wireless channels, we assume that the channel gain of each channel is determined by three components:
\begin{itemize}
    \item \emph{Path-Loss}: modeled by the short-range outdoor model ITU-1411. 
    \item \emph{Shadowing}: modeled by the log-normal distribution with 8dB standard deviation.
    \item \emph{Fast Fading}: modeled by Rayleigh fading with i.i.d circular Gaussian distribution of unit variance.
\end{itemize}

We collect $N^2$ channel gains for each layout into a $N^2$-dimensional vector as the input $\mathbf{p}$. Similar to Section~\ref{sec:result_I}, we utilize the same neural network specification in $\Theta_s$ and $\Theta_t$ for both the feature learning stage and the optimization stage (since $\mathbf{x}_s$ and $\mathbf{x}_t$ have the same dimension as the number of links $N$). The same overall neural network specification is used by all the competing methods, as summarized in Table~\ref{tab:d2d_neuralnet}. With $N=10$, the total numbers of trainable parameters for all three stages in the neural network computation flows (including both weights and biases) are as follows:
\begin{itemize}
    \item \emph{Feature Learning Stage ($\Theta_s$ or $\Theta_t$)}: 52900 parameters;
    \item \emph{Optimization Stage ($\Theta_s$ or $\Theta_t$)}: 5070 parameters;
    \item \emph{Reconstruction Stage (only for $\Theta_s$ trained by our proposed approach)}: 40300 parameters.
\end{itemize}
As the optimization stage only has a small number of parameters, training models on $\mathcal{T}$ via transfer learning requires little data. 

\begin{table}[t]
\caption{D2D Transfer Learning Neural Network Architecture \mbox{($N$: number of links)}}
\begin{center}
\begin{tabular}{|l|c|c|}
\hline
\topstrut\textbf{Stages} & \textbf{Layers} & \textbf{Number of Neurons}\\
\hline
\multirow{3}*{\shortstack[l]{Feature Learning\\ Stage ($\Theta_s$ or $\Theta_t$)}}
& \topstrut 1st & $1.5N^2$ \\ \cline{2-3}
& \topstrut 2nd & $1.5N^2$ \\ \cline{2-3}
& \topstrut Feature Layer & $N^2$ \\ 
\hline
\multirow{3}*{\shortstack[l]{Optimization \\ Stage ($\Theta_s$ or $\Theta_t$)}} 
& \topstrut 1st & $4N$ \\ \cline{2-3}
& \topstrut 2nd & $2N$ \\ \cline{2-3}
& \topstrut Output Layer & $N$ \\
\hline
\multirow{2}*{\shortstack[l]{Reconstruction \\ Stage (only $\Theta_s$)}} 
& \topstrut 1st & $2N^2$ \\ \cline{2-3}
& \topstrut Reconstruct Layer & $N^2$ \\ 
\hline
\end{tabular}
\label{tab:d2d_neuralnet}
\end{center}
\end{table}

To train each neural network, we formulate the (unsupervised-learning) loss functions $\mathcal{L_S}$ and $\mathcal{L_T}$ directly based on the task utility values of both tasks as follows:
\begin{align}
    \mathcal{L_S} =&\: -u_s(\mathbf{x}_s) \:,\\
    \mathcal{L_T} =&\: -u_t(\mathbf{x}_t) \:,
\end{align}
with $u_s(\mathbf{x}_s)$ and $u_t(\mathbf{x}_t)$ defined as (\ref{equ:sum_rate}) and (\ref{equ:min_rate}) respectively. Note that we define the loss functions by negating the utility functions, such that the utility functions are maximized while the loss functions are minimized. Furthermore, to train $\mathcal{F}_{\Theta_s}$ with our approach, we use $\alpha=3$ in (\ref{equ:loss}) for the loss $\mathcal{L}$, which provides the best performance in the process of hyper-parameter tuning.

In terms of the data used for training under all competing methods, we utilize the data set specification listed in Table~\ref{tab:d2d_dataset}. Each sample of a D2D wireless network is generated according to the wireless network simulation settings mentioned earlier. We note that the data set sizes in Table~\ref{tab:d2d_dataset} are smaller than the number of neural network trainable parameters, especially for the training data on $\mathcal{T}$. Similar to Section~\ref{sec:result_I}, we select these small data sets to illustrate that new target tasks can be adapted on-the-fly with minimal training overhead, as well as to show that the proposed approach is effective in knowledge transfer and is robust to over-fitting. We use relatively large validation sets to ensure accurate early stopping when training the model under each method. However, we may not have sufficient data for validation when training the model on $\mathcal{T}$ in realistic scenarios. Therefore, in Section~\ref{sec:result_II_II}, we also include simulation results where no early stopping is performed during target task training under each method. 

\ifOneCol
    \begin{table}[t]
    \caption{D2D Wireless Networks Data Set Specifications}
    \begin{center}
    \begin{tabular}{|c|c|c|}
    \hline
    \textbf{Task} & \textbf{Training Set Samples} & \textbf{Validation Set Samples}\\
    \hline
    $\mathcal{S}$
    &  $5\times 10^5$ & 5000 \\ 
    \hline
    $\mathcal{T}$ 
    &  1000 & 5000* \\ 
    \hline
    \end{tabular}
    \label{tab:d2d_dataset}
    \end{center}
    {\footnotesize $^*$
    A large validation set is used to ensure accurate early stopping. May not be available in a realistic scenario in which the target task data is limited.}
    \end{table}
\else
    \begin{table}[t]
    \caption{D2D Wireless Networks Data Set Specifications}
    \begin{center}
    \begin{tabular}{|c|c|c|}
    \hline
    \topstrut \textbf{Task} & \textbf{Training Set Samples} & \textbf{Validation Set Samples}\\
    \hline
    \topstrut $\mathcal{S}$
    &  $5\times 10^5$ & 5000 \\ 
    \hline
    \topstrut $\mathcal{T}$ 
    &  1000 & 5000* \\ 
    \hline
    \end{tabular}
    \label{tab:d2d_dataset}
    \end{center}
    {\footnotesize $^*$
    A large validation set is used to ensure accurate early stopping. May not be available in a realistic scenario in which the target task data is limited.}
    \end{table}
\fi

For the test data set (on which both utilities $u_s$ and $u_t$ are evaluated), we generate 2000 new samples of D2D wireless networks to obtain performance statistics over all the methods. 

\subsubsection{Transfer Learning Performances}\label{sec:result_II_II}
Besides the two neural network based benchmarks, we also include the following traditional mathematical optimization algorithms serving as performance upper-bound baselines (with the cost of having much higher computational complexities):
\begin{itemize}
    \item \emph{Geometric Programming (GP)}\cite{gp}: mathematical algorithm for solving the min-rate optimization.
    \item \emph{Fractional Programming (FP)}\cite{fplinq}: mathematical algorithm for solving the sum-rate optimization.
\end{itemize}
We train and evaluate each neural network based method under the transfer learning direction: Sum Rate~$\mathcal{S}\to$~Min Rate~$\mathcal{T}$, while only evaluating FP on $\mathcal{S}$ and GP on $\mathcal{T}$.

We present both $\mathcal{S}$ and $\mathcal{T}$ performances, averaged over all 2000 test wireless networks, in Table~\ref{tab:d2d_results}. As a first observation from the results, the conventional transfer learning approach performs worse than even the regular learning method, indicating that this D2D power control optimization application indeed poses as a challenging problem for transfer learning. Therefore, the proposed approach is much needed for achieving knowledge transfer on such general mathematical optimization problems. We then focus on the results \emph{with early stopping}, obtaining which requires additional data reserved as the validation set on $\mathcal{T}$. Shown by the numerical results, when the training data on the target task is limited (1000 samples as in Table~\ref{tab:d2d_dataset}), the transfer learning-with-reconstruction approach achieves the best target-task performance among the neural network based methods, with a 11\% improvement over the regular learning approach, and a 17\% improvement over the conventional transfer learning approach. The proposed approach achieves these improvements while sacrificing minimal source-task performance as the trade-off: a 1\% reduction on the sum-rate results as compared to both neural network based benchmarks. This slight loss of the performance on $\mathcal{T}$ is expected since our approach utilizes the training loss as in (\ref{equ:loss}) that does not exclusively target at optimizing the source-task utility. 

\ifOneCol
    \begin{table*}[t]
    \caption{D2D Wireless Networks Transfer Learning Performances}
    \begin{center}
    \begin{tabular}{|c|c|c|c|}
    \hline
    \topstrut \textbf{Task} & \textbf{Method} & \makecell{\shortstack[c]{\topstrut\textbf{Result with} \\ \textbf{Early Stopping (Mbps)}}} & \makecell{\shortstack[c]{\topstrut\textbf{Result without} \\ \textbf{Early Stopping (Mbps)}}}\\
    \hline
    \multirow{4}*{$\mathcal{S}$: Sum-Rate}
    & \topstrut Regular Learning & \multirow{2}*{155.69} & \multirow{4}*{N/A} \\\cline{2-2}
    & \topstrut Conventional Transfer Learning & & \\\cline{2-3}
    & \topstrut Transfer Learning with Reconstruction & 153.96 &  \\\cline{2-3}
    & \topstrut FP & 157.45 & \\
    \hhline{|=|=|=|=|}
    \multirow{4}*{$\mathcal{T}$: Min-Rate}
    & \topstrut Regular Learning & 5.39 & 5.32 \\\cline{2-4}
    & \topstrut Conventional Transfer Learning & 5.13 & 4.80 \\\cline{2-4}
    & \topstrut Transfer Learning with Reconstruction & 6.00 & 6.00 \\\cline{2-4}
    & \topstrut GP & 9.16 & N/A \\
    \hline
    \end{tabular}
    \label{tab:d2d_results}
    \end{center}
    \end{table*}
\else
    \begin{table*}[t]
    \caption{D2D Wireless Networks Transfer Learning Performances}
    \begin{center}
    \begin{tabular}{|c|c|c|c|}
    \hline
    \topstrut \textbf{Task} & \textbf{Method} & \textbf{Result with Early Stopping (Mbps)} & \textbf{Result without Early Stopping (Mbps)}\\
    \hline
    \multirow{4}*{$\mathcal{S}$: Sum-Rate}
    & \topstrut Regular Learning & \multirow{2}*{155.69} & \multirow{4}*{N/A} \\\cline{2-2}
    & \topstrut Conventional Transfer Learning & & \\\cline{2-3}
    & \topstrut Transfer Learning with Reconstruction & 153.96 &  \\\cline{2-3}
    & \topstrut FP & 157.45 & \\
    \hhline{|=|=|=|=|}
    \multirow{4}*{$\mathcal{T}$: Min-Rate}
    & \topstrut Regular Learning & 5.39 & 5.32 \\\cline{2-4}
    & \topstrut Conventional Transfer Learning & 5.13 & 4.80 \\\cline{2-4}
    & \topstrut Transfer Learning with Reconstruction & 6.00 & 6.00 \\\cline{2-4}
    & \topstrut GP & 9.16 & N/A \\
    \hline
    \end{tabular}
    \label{tab:d2d_results}
    \end{center}
    \end{table*}
\fi

\subsubsection{Learning Dynamics and Over-fitting}\label{sec:result_II_III}
\begin{figure*}[t]
    \centering
    \includegraphics[width=0.9\textwidth]{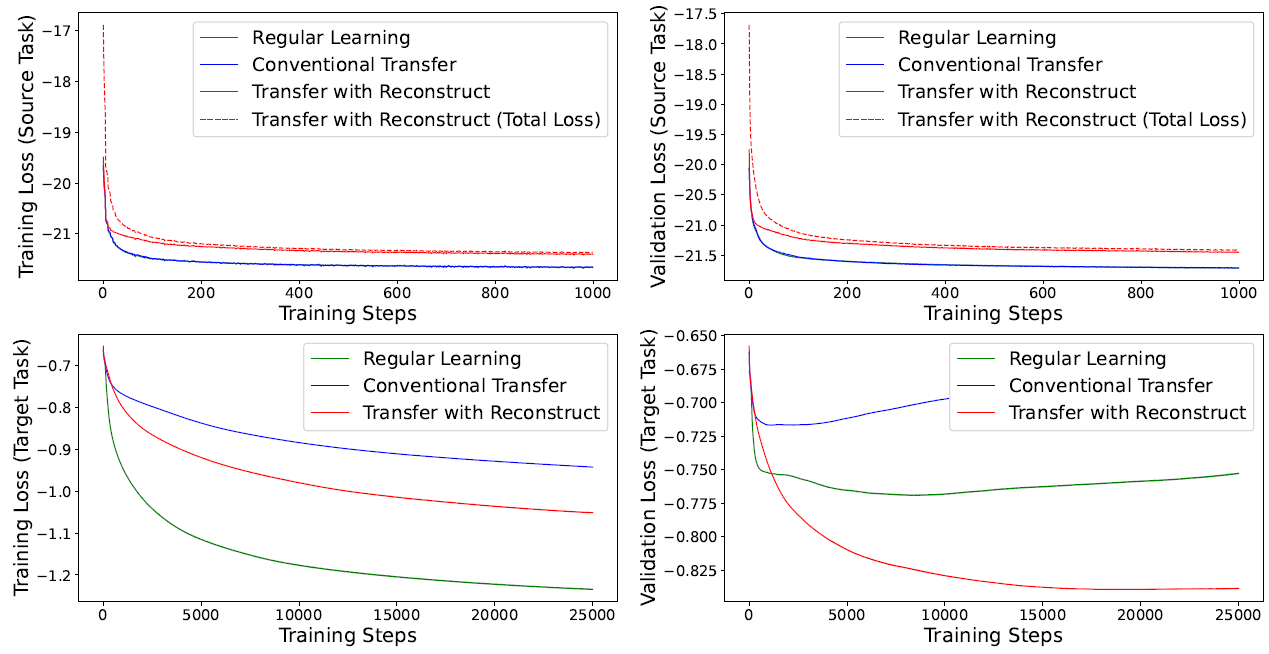}
    \caption{Training Curves for Transfer Learning in D2D Wireless Network Optimizations (the top two figures provide training and validation curves for the source task; the bottom two figures provide training and validation curves for the target task).}
    \label{fig:traincurve}
\end{figure*}

To understand the training dynamics that lead to the presented results and to visualize if and how the over-fitting occurs for each method, we provide the training curves on $\mathcal{S}$ and $\mathcal{T}$ for all the methods in Fig.~\ref{fig:traincurve}. Note that for the proposed approach, we have plotted two losses on $\mathcal{S}$: the source task based loss shown by the solid line ($\mathcal{L_S}$ in (\ref{equ:loss})), and the total loss shown by the dotted line ($\mathcal{L}$ in (\ref{equ:loss})). As evident by the validation curves on $\mathcal{T}$ (in the bottom-right figure), while both the conventional transfer learning and the regular learning approaches plateau early in validation loss and then regress noticeably due to over-fitting, our approach enables the model to learn at a much more sustainable pace from the very limited training data set of 1000 samples, without any noticeable over-fitting. 

The effects of over-fitting are shown more clearly from the simulation results when no early stopping is performed during training on $\mathcal{T}$. Examining again Table~\ref{tab:d2d_results}, as shown by the results \emph{without early stopping}, our approach maintains its performance on $u_t$ (indicating the model does not over-fit throughout training) and enjoys larger margins on $\mathcal{T}$, with 13\% and 25\% improvements over the regular learning and conventional transfer learning approach respectively. As this set of results are achieved without needing a large validation set on $\mathcal{T}$, the performance comparison results are more relevant to practical implementations.

\subsection{Transfer Learning on MISO Downlink Wireless Networks}\label{sec:result_III}
\subsubsection{Simulation Settings}\label{sec:result_III_I}
For the MISO downlink wireless network application, we simulate each wireless network within a 3-dimensional confined region with dimensions of 100m$\times$100m$\times$50m. We set up $M=3$ BSs, located at the fixed locations at the top level of the region: (0m, 0m, 50m), (0m, 100m, 50m), (100m, 0m, 50m). Each BS is equipped with $K=16$ antennas, arranged in a 4$\times$4 2-dimensional array in parallel to the x-y plane. These BS configurations stay the same across all instances of MISO wireless networks generated. In each MISO network, the UE is located at an unknown location uniformly generated within the planar region (50$\pm$40m, 50$\pm$40m, 0m). We assume a maximum transmission power level of $P_{BS}=40$ dBm for each BS and $P_{UE}=30$ dBm for the UE. We also assume a background noise level at -150 dBm/Hz.

For wireless channel specifications, we use the following pathloss model in the decibel (dB) scale to model the pathloss components of the channels ($\rho(d_m)$ in (\ref{equ:misochannel})):
\begin{align}
    \rho(d_m)_{dB} = -32.6-36.7\log_{10}(d_m).
\end{align}
Without loss of generality, we assume antenna spacing $\delta$ and signal wavelength $\lambda$ values such that $\frac{2\pi\delta}{\lambda}=1$. The Rician factor $\epsilon=10$ is used. For channel estimation, the UE transmits $T=4$ uplink pilots. To receive the four uplink pilots, each BS employs four sensing vectors. The uplink pilots and the digital sensing vectors are all randomly generated by sampling from circularly-symmetric complex normal distribution, and then normalize to the proper power levels. Specifically, the pilots are generated as follows:
\begin{align}
    c^{t'}\sim&\:\mathcal{CN}(0, \mathbf{I}) \\
    c^t=&\:\sqrt{P_{\text{UE}}}\frac{c^{t'}}{|c^{t'}|} \quad \forall t\:,
\end{align}
while the sensing vectors are generated as follows:
\begin{align}
    \mathbf{v}_m'\sim&\:\mathcal{CN}(0, \mathbf{I}) \\
    \mathbf{v}_m=&\:\frac{\mathbf{v}_m'}{||\mathbf{v}_m'||_2} \quad \forall m,t\:.
\end{align}
As mentioned in Section~\ref{sec:prob_III}, we use a fixed set of pilots $\{c^t\}_{t\in\{1\dots T\}}$ and sensing vectors over all BSs $\{\mathbf{v}_m\}_{m\in\{1\dots M\}}$ over all the training and testing MISO wireless network instances, and therefore the models can implicitly learn the knowledge about these pilots and sensing vectors and further estimate the varying wireless channels $\mathbf{H}$ by using only the inputs $\{r_m^t\}_{m\in\{1\dots M\},t\in\{1\dots T\}}$ as in (\ref{equ:pilotmeasures}).

As specified in (\ref{equ:miso_ci}), we select the three sets of geometric parameters as the common information: the set of distances $\{d_m\}_{m\in\{1\dots M\}}$, the set of azimuth AoAs $\{\theta_m\}_{m\in\{1\dots M\}}$, and the set of elevation AoAs $\{\phi_m\}_{m\in\{1\dots M\}}$. In the actual implementation, we make the observation that in (\ref{equ:misochannel_LOS2}), $\{\theta_m\}$ and $\{\phi_m\}$ occur in the terms $\sin(\theta_m)\cos(\phi_m)$ and $\cos(\theta_m)\cos(\phi_m)$. Therefore, in our simulations, we collect the following values as the common information:
\ifOneCol
    \begin{align}\label{equ:miso_ci2}
        \mathcal{I}(\mathbf{p})\leftarrow
        \{d_m, \sin(\theta_m)\cos(\phi_m),\cos(\theta_m)\cos(\phi_m)\}_{m\in\{1\dots M\}}.
    \end{align}
\else
    \begin{multline}\label{equ:miso_ci2}
        \mathcal{I}(\mathbf{p})\leftarrow\\
        \{d_m, \sin(\theta_m)\cos(\phi_m),\cos(\theta_m)\cos(\phi_m)\}_{m\in\{1\dots M\}}.
    \end{multline}
\fi
Although the common information in (\ref{equ:miso_ci2}) is not as concise as in (\ref{equ:miso_ci}), the simulation results suggest that using (\ref{equ:miso_ci2}) as the common information in our proposed approach is still effective in obtaining highly competitive transfer learning performances.

We provide in Table~\ref{tab:miso_dataset} and Table~\ref{tab:miso_neuralnet} the specifications for data sets and the neural network architecture used by all the competing methods. We reiterate that for the MISO downlink transmission application, $\mathbf{x}_s$ and $\mathbf{x}_t$ are solutions for the optimal beamformers and the UE location estimation respectively. Therefore, unlike the previous two applications in Section~\ref{sec:result_I} and \ref{sec:result_II}, the optimization variables $\mathbf{x}_s$ and $\mathbf{x}_t$ have different dimensions: $\mathbf{x}_s\in\mathbb{R}^{2MK}$ contain the real and imaginary parts of the $M$ beamformers $\{\mathbf{b}_m\}_{m\in\{1\dots M\}}$; while $\mathbf{x}_t\in\mathbb{R}^2$ contain the location coordinate estimation ($\hat{x}_{\text{UE}}$, $\hat{y}_{\text{UE}}$) for the UE. Correspondingly, we construct the optimization stages in $\Theta_s$ and $\Theta_t$ using different specifications as shown in Table~\ref{tab:miso_neuralnet}. We also note that for the common information as in (\ref{equ:miso_ci2}), three values are collected for each of the $M$ BSs. Therefore, the dimension of the common information, which is also the output dimension for the reconstruction stage, is $3M=9$. Specifically, the numbers of trainable parameters for all three stages in the neural network computation flows (including weights and biases), in $\Theta_s$ and $\Theta_t$, are as follows:
\begin{itemize}
    \item \emph{Feature Learning Stage ($\Theta_s$ or $\Theta_t$)}: 64150 parameters;
    \item \emph{Optimization Stage in $\Theta_s$}: 29896 parameters;
    \item \emph{Optimization Stage in $\Theta_t$}: 36702 parameters;
    \item \emph{Reconstruction Stage (only for $\Theta_s$ trained by our proposed approach)}: 8059 parameters
\end{itemize}
As shown above, with carefully crafted low-dimensional common information, the number of extra parameters introduced by our approach as the reconstruction stage is relatively low in the source-task training. 

\ifOneCol
    \begin{table}[t]
    \caption{MISO Wireless Networks Data Set Specifications}
    \begin{center}
    \begin{tabular}{|c|c|c|}
    \hline
    \textbf{Task} & \textbf{Training Set Samples} & \textbf{Validation Set Samples}\\
    \hline
    $\mathcal{S}$
    &  $1\times 10^5$ & 5000 \\ 
    \hline
    $\mathcal{T}$ 
    &  500 & 5000* \\ 
    \hline
    \end{tabular}
    \label{tab:miso_dataset}
    \end{center}
    {\footnotesize $^*$
    A large validation set is used to ensure accurate early stopping. May not be available in a realistic scenario in which the target task data is limited. }
    \end{table}
\else
    \begin{table}[t]
    \caption{MISO Wireless Networks Data Set Specifications}
    \begin{center}
    \begin{tabular}{|c|c|c|}
    \hline
    \topstrut \textbf{Task} & \textbf{Training Set Samples} & \textbf{Validation Set Samples}\\
    \hline
    \topstrut $\mathcal{S}$
    &  $1\times 10^5$ & 5000 \\ 
    \hline
    \topstrut $\mathcal{T}$ 
    &  500 & 5000* \\ 
    \hline
    \end{tabular}
    \label{tab:miso_dataset}
    \end{center}
    {\footnotesize $^*$
    A large validation set is used to ensure accurate early stopping. May not be available in a realistic scenario in which the target task data is limited. }
    \end{table}
\fi

\ifOneCol
    \begin{table}[t]
    \caption{MISO Transfer Learning Neural Network Architecture ($M$: Number of BSs; $N$: Number of Antennas per BS)}
    \begin{center}
    \begin{tabular}{|l|c|c|}
    \hline
    \textbf{Stages} & \textbf{Layers} & \textbf{Number of Neurons}\\
    \hline
    \multirow{4}*{\shortstack[l]{Feature Learning\\ Stage ($\Theta_s$ or $\Theta_t$)}}
    & 1st & 150 \\ \cline{2-3}
    & 2nd & 150 \\ \cline{2-3}
    & 3rd & 150 \\ \cline{2-3}
    & Feature Layer & 100 \\ 
    \hline
    \multirow{3}*{\shortstack[l]{Optimization \\ Stage for $\Theta_s$}} 
    & 1st & 100 \\ \cline{2-3}
    & 2nd & 100 \\ \cline{2-3}
    & Output Layer & $2MK$ \\
    \hline
    \multirow{5}*{\shortstack[l]{Optimization \\ Stage for $\Theta_t$}} 
    & 1st & 125 \\ \cline{2-3}
    & 2nd & 100 \\ \cline{2-3}
    & 3rd & 75 \\ \cline{2-3}
    & 4th & 50 \\ \cline{2-3}
    & Output Layer & 2 \\
    \hline
    \multirow{3}*{\shortstack[l]{Reconstruction \\ Stage (only $\Theta_s$)}} 
    & 1st & 50 \\ \cline{2-3}
    & 2nd & 50 \\ \cline{2-3}
    & Reconstruct Layer & $3M$ \\ 
    \hline
    \end{tabular}
    \label{tab:miso_neuralnet}
    \end{center}
    \end{table}
\else
    \begin{table}[t]
    \caption{MISO Transfer Learning Neural Network Architecture \mbox{($M$: Number of BSs;} $N$: Number of Antennas per BS)}
    \begin{center}
    \begin{tabular}{|l|c|c|}
    \hline
    \topstrut\textbf{Stages} & \textbf{Layers} & \textbf{Number of Neurons}\\
    \hline
    \multirow{4}*{\shortstack[l]{Feature Learning\\ Stage ($\Theta_s$ or $\Theta_t$)}}
    & \topstrut 1st & 150 \\ \cline{2-3}
    & \topstrut 2nd & 150 \\ \cline{2-3}
    & \topstrut 3rd & 150 \\ \cline{2-3}
    & \topstrut Feature Layer & 100 \\ 
    \hline
    \multirow{3}*{\shortstack[l]{Optimization \\ Stage for $\Theta_s$}} 
    & \topstrut 1st & 100 \\ \cline{2-3}
    & \topstrut 2nd & 100 \\ \cline{2-3}
    & \topstrut Output Layer & $2MK$ \\
    \hline
    \multirow{5}*{\shortstack[l]{Optimization \\ Stage for $\Theta_t$}} 
    & \topstrut 1st & 125 \\ \cline{2-3}
    & \topstrut 2nd & 100 \\ \cline{2-3}
    & \topstrut 3rd & 75 \\ \cline{2-3}
    & \topstrut 4th & 50 \\ \cline{2-3}
    & \topstrut Output Layer & 2 \\
    \hline
    \multirow{3}*{\shortstack[l]{Reconstruction \\ Stage (only $\Theta_s$)}} 
    & \topstrut 1st & 50 \\ \cline{2-3}
    & \topstrut 2nd & 50 \\ \cline{2-3}
    & \topstrut Reconstruct Layer & $3M$ \\ 
    \hline
    \end{tabular}
    \label{tab:miso_neuralnet}
    \end{center}
    \end{table}
\fi

In terms of the selection for $\mathcal{L_S}$, unlike in Section~\ref{sec:result_II}, we have discovered through simulations that training models by formulating the source-task loss function as the utility value $u_s$ (as in (\ref{equ:beamform_utility})) does not lead to as strong performances as compared to training with supervised learning targets. For the source task $\mathcal{T}$ on downlink beamforming, we have the training targets being the \emph{perfect beamformers}, which are the beamformers perfectly aligned with the wireless channels from each BS to the UE, which we denote by $\mathbf{B}^*=\{\mathbf{b}_m^*\}_{m\in\{1\to M\}}$, obtained as follows:
\begin{align}\label{equ:miso_perfect}
    \mathbf{b}_m^*=\frac{\mathbf{h}_m}{||\mathbf{h}_m||_2}\quad\forall m\:,
\end{align}
where $\mathbf{h}_m$ is the set of actual channels from the $m$-th BS to the UE, as described in Section~\ref{sec:prob_III}. With $\mathbf{B}^*$ being the target labels, we formulate the source-task loss function $\mathcal{L_S}$ as the squared-error loss (which is a popular choice of loss function for regression tasks):
\begin{align}
    \mathcal{L_S}=\sum_{m=1}^M||\mathbf{b}_m^\text{perfect}-\mathbf{x}_s^m||_2^2 \:,
\end{align}
where $\mathbf{x}_s^m$ is the optimized beamformers for the $m$-th BS as defined in (\ref{equ:mimo_xs}). With the target task $\mathcal{T}$ being the downlink localization task, it is naturally formulated as a regression task with the true UE location being the target. Correspondingly, we use the squared-error loss function $\mathcal{L_T}$:
\begin{align}
    \mathcal{L_T}=(x_\text{UE}-\hat{x}_{\text{UE}})^2+(y_\text{UE}-\hat{y}_{\text{UE}})^2 \:,
\end{align}
with $\hat{x}_{\text{UE}}$ and $\hat{y}_{\text{UE}}$ being the UE location estimations as the model's output $\mathbf{x}_t$, as specified in (\ref{equ:mimo_xt}). To train $\mathcal{F}_{\Theta_s}$ with our approach, we use $\alpha=4$ in (\ref{equ:loss}) for the loss $\mathcal{L}$, which provides the best performance in the process of hyper-parameter tuning.

\subsubsection{Transfer Learning Performances}\label{sec:result_III_II}
For the downlink beamforming task $\mathcal{S}$, we provide several baselines in addition to the two neural network based benchmarks, as follows:
\begin{itemize}
    \item \emph{Perfect Beamformers}: assuming that accurate knowledge of all wireless channels $\mathbf{H}$ is available, the beamformers are designed to align with the channels from each BS to the UE, computed as in (\ref{equ:miso_perfect}).
    \item \emph{Random Beamformers}:
    randomly generate beamformers by firstly generate entries from circularly-symmetric complex normal distribution and then apply normalization to ensure unit beamformer power.
\end{itemize}

We test all the methods over 2000 newly generated testing MISO wireless networks, and present results on $\mathcal{S}$ and $\mathcal{T}$ in Table~\ref{tab:miso_results}, where we report the average of the SNR values as $u_s(\mathbf{x}_s)$ in (\ref{equ:beamform_utility}) and the average of the localization errors as $u_t(\mathbf{x}_t)$ in (\ref{equ:localization_utility}). Similar to Section~\ref{sec:result_II_II}, we have also included in Table~\ref{tab:miso_results} the target task performances by each method 
when no early stopping is performed during training. Furthermore, we also present in Fig.~\ref{fig:localization_cdf} the CDF curves over 2000 testing MISO wireless networks for the localization errors $u_t$. 

\ifOneCol
    \begin{table*}[t]
    \caption{MISO Downlink Wireless Networks Transfer Learning Performances}
    \begin{center}
    \begin{tabular}{|c|c|c|c|}
    \hline
    \topstrut \textbf{Task} & \textbf{Method} & \makecell{\shortstack[c]{\topstrut \textbf{Result with}\\\textbf{Early Stopping}}} & \makecell{\shortstack[c]{\topstrut\textbf{Result without}\\\textbf{Early Stopping}}}\\
    \hline
    \multirow{5}*{$\mathcal{S}$: Beamforming}
    & \topstrut Regular Learning & \multirow{2}*{36.47dB} & \multirow{5}*{N/A} \\\cline{2-2}
    & \topstrut Conventional Transfer Learning & & \\\cline{2-3}
    & \topstrut Transfer Learning with Reconstruction & 36.43dB &  \\\cline{2-3}
    & \topstrut Random Beamformers & 23.79dB & \\\cline{2-3}
    & \topstrut Perfect Beamformers & 36.83dB & \\
    \hhline{|=|=|=|=|}
    \multirow{3}*{$\mathcal{T}$: Localization}
    & \topstrut Regular Learning & 6.56m & 7.79m \\\cline{2-4}
    & \topstrut Conventional Transfer Learning & 6.06m & 6.07m \\\cline{2-4}
    & \topstrut Transfer Learning with Reconstruction & 5.59m & 5.60m \\
    \hline
    \end{tabular}
    \label{tab:miso_results}
    \end{center}
    \end{table*}
\else
    \begin{table*}[t]
    \caption{MISO Downlink Wireless Networks Transfer Learning Performances}
    \begin{center}
    \begin{tabular}{|c|c|c|c|}
    \hline
    \topstrut \textbf{Task} & \textbf{Method} & \textbf{Result with Early Stopping} & \textbf{Result without Early Stopping}\\
    \hline
    \multirow{5}*{$\mathcal{S}$: Beamforming}
    & \topstrut Regular Learning & \multirow{2}*{36.47dB} & \multirow{5}*{N/A} \\\cline{2-2}
    & \topstrut Conventional Transfer Learning & & \\\cline{2-3}
    & \topstrut Transfer Learning with Reconstruction & 36.43dB &  \\\cline{2-3}
    & \topstrut Random Beamformers & 23.79dB & \\\cline{2-3}
    & \topstrut Perfect Beamformers & 36.83dB & \\
    \hhline{|=|=|=|=|}
    \multirow{3}*{$\mathcal{T}$: Localization}
    & \topstrut Regular Learning & 6.56m & 7.79m \\\cline{2-4}
    & \topstrut Conventional Transfer Learning & 6.06m & 6.07m \\\cline{2-4}
    & \topstrut Transfer Learning with Reconstruction & 5.59m & 5.60m \\
    \hline
    \end{tabular}
    \label{tab:miso_results}
    \end{center}
    \end{table*}
\fi

\begin{figure*}[t]
    \centering
    \includegraphics[width=0.85\textwidth]{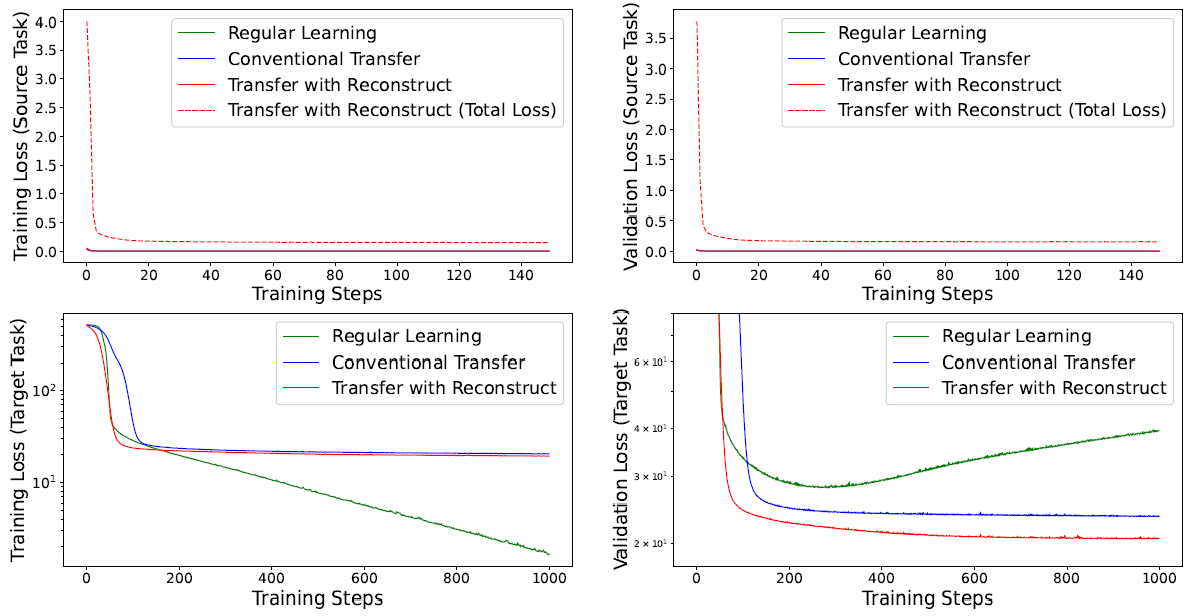}
    \caption{Training Curves for Transfer Learning in MISO Wireless Network Optimizations (the top two figures provide training and validation curves for the source task; the bottom two figures provide training and validation curves for the target task).}
    \label{fig:miso_train}
\end{figure*}

\ifOneCol
    \begin{figure}[t]
        \centering
        \includegraphics[width=0.8\textwidth]{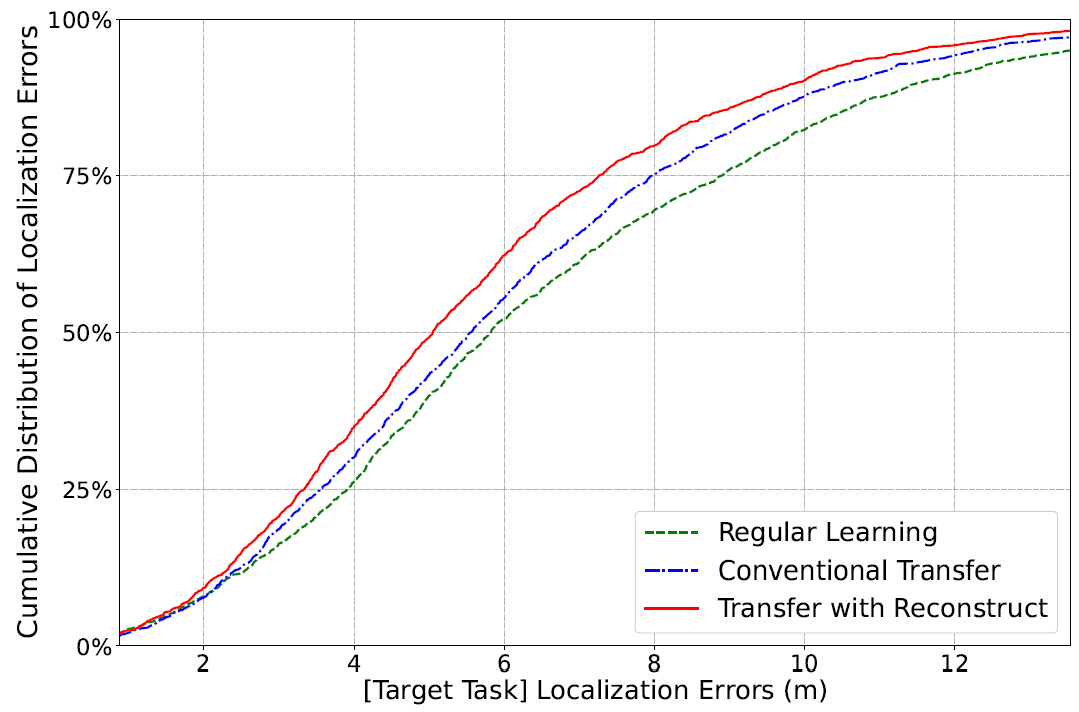}
        \caption{CDF for MISO network downlink localization errors (the more to the left the curve locates, the lower the localization errors are).}
        \label{fig:localization_cdf}
    \end{figure}
\else
    \begin{figure}[t]
        \centering
        \includegraphics[width=0.475\textwidth]{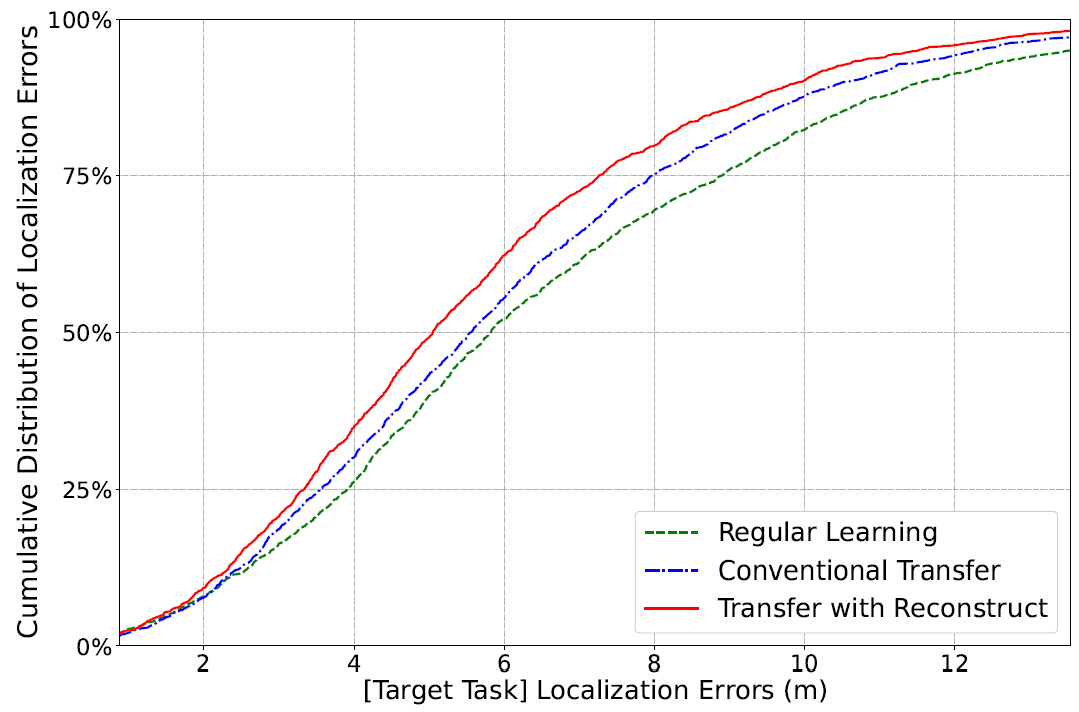}
        \caption{CDF for MISO network downlink localization errors (the more to the left the curve locates, the lower the localization errors are).}
        \label{fig:localization_cdf}
    \end{figure}
\fi

\begin{figure*}[t]
    \centering
    \subfigure[Localization on MISO network sample \# 1]{\includegraphics[width=0.4\textwidth]{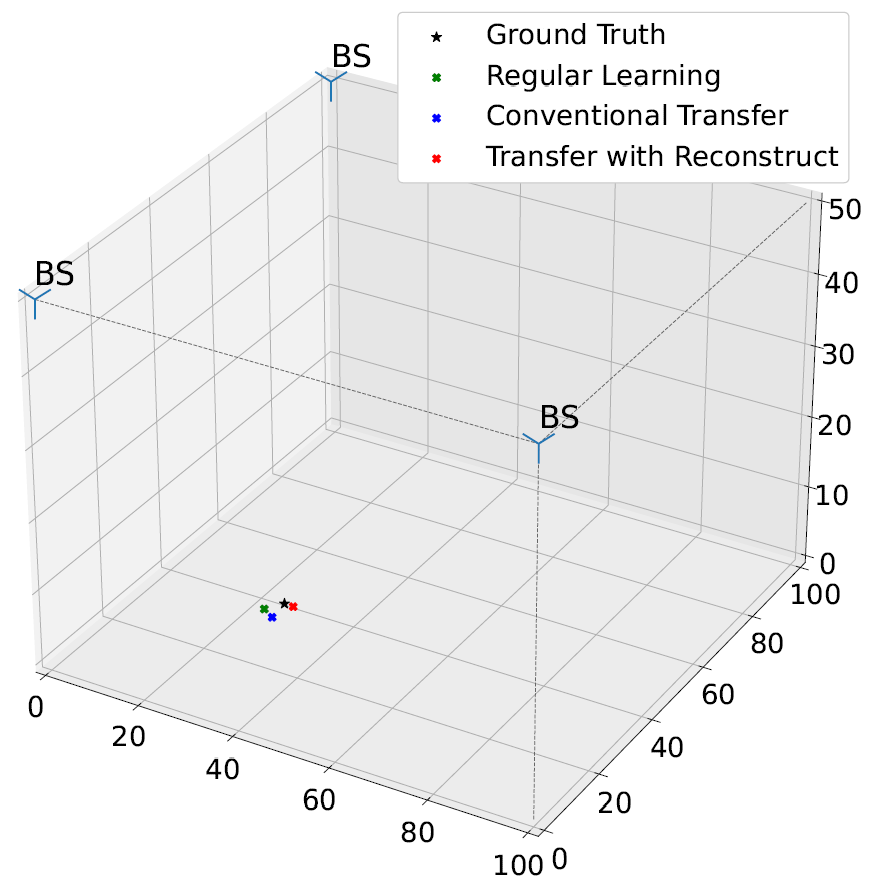}}
    \subfigure[Localization on MISO network sample \# 2]{\includegraphics[width=0.4\textwidth]{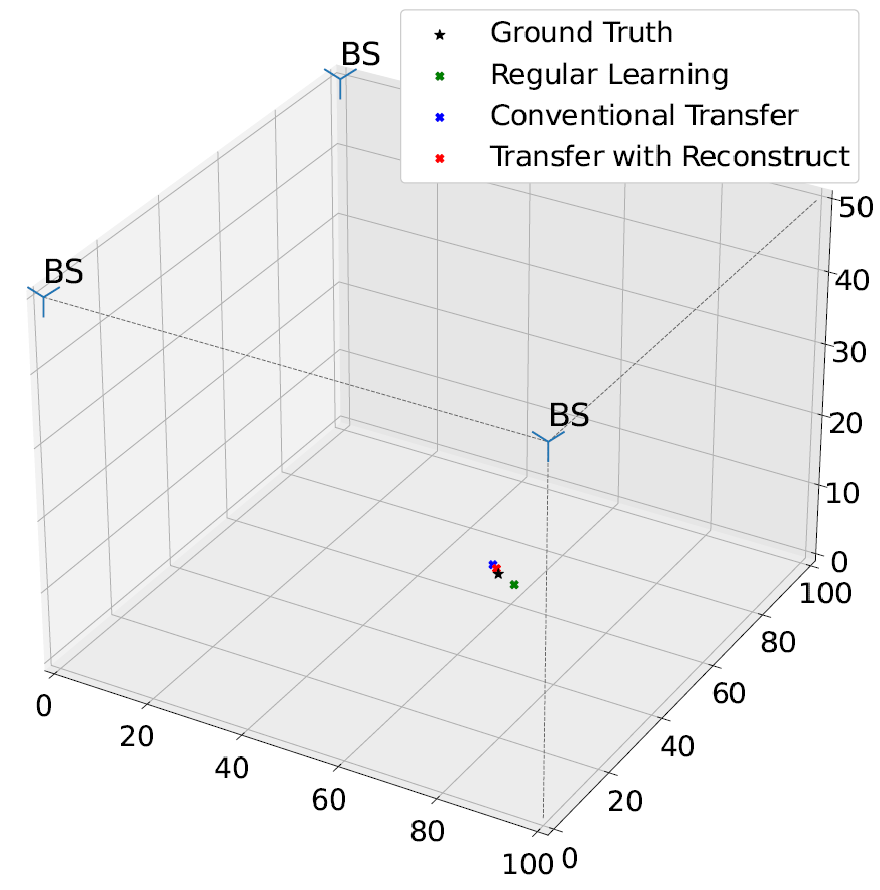}}
    \caption{Visualization of localization results in two randomly selected testing samples of MISO networks.}
    \label{fig:localization_sample}
\end{figure*}

As suggested by the performance statistics in Table~\ref{tab:miso_results}, the proposed transfer learning with reconstruction loss approach achieves a better target task localization performance, with a 15\% improvement over the regular learning method and a 8\% improvement over the convetional transfer learning method. To visualize the performance margins, we provide in Fig.~\ref{fig:localization_sample} visualizations of the localization results by all the methods in two testing MISO wireless network samples. 

\subsubsection{Learning Dynamics and Over-fitting}
To show how the proposed approach excels, we provide in Fig.~\ref{fig:miso_train} the training and validation curves on both $\mathcal{S}$ and $\mathcal{T}$ by all the methods. Under limited target task training data (500 samples as specified in Table~\ref{tab:miso_dataset}), as clearly shown by the validation loss curves in Fig.~\ref{fig:miso_train}, the model trained with the proposed approach shows little overfitting and maintains a much more sustainable and effective learning progress compared to the other two methods. These observed target-task learning dynamics validate the fact that the proposed approach effectively addresses the over-fitting issue in transfer learning.

The effects from over-fitting are further reflected in the comparison between the localization performances with and without early stopping under each method, as shown in Table~\ref{tab:miso_results}. Similar to the previous two applications, the proposed method achieves the better target task performances with minimal performance tradeoff on the source task. 

\subsubsection{Concise Representation of Common Information}
We would also like to emphasize that these performance gains are achieved with the common information in a concise representation tailored specifically to the tasks as in (\ref{equ:miso_ci}) or (\ref{equ:miso_ci2}). Benefited from such low-dimensional common information, the additional reconstruction stage, which is introduced into $\Theta_s$ by the proposed approach only during training for $\mathcal{S}$, also enjoys low parameter and computational complexities relative to the feature learning stage and the optimization stage of the model. This application illustrates the effectiveness of the proposed method when the domain knowledge is applied on selecting the task-specific common information.

\section{Conclusion}\label{sec:conclusion}
Transfer learning has great potential and wide applicability in general mathematical optimization problems. However, when using neural networks to learn such optimization mappings, it is challenging to learn or to identify the transferable features within the neural network computation flows, due to the lack of structures in the inputs and in the computation flows. This paper proposes a novel transfer learning approach for learning general and transferable features at specified locations within neural networks. We first establish the concept of common information in correlated tasks, the choice of which can be generic or problem-specific. We then introduce a reconstruction stage in the neural network model starting from a pre-specified hidden layer, i.e., the feature layer. By enforcing the reconstruction of the common information based on the features from the feature layer, these learned features are generally descriptive of all the correlated tasks, and therefore can be transferred among multiple task optimizations. Simulation results on the MNIST classification problem and two wireless network utility optimization problems suggest that the proposed approach consistently outperforms the conventional transfer learning method and is robust against over-fitting under limited training data. We hope this work could help open up further exploration on bridging together transfer learning and general mathematical optimizations.

\bibliographystyle{IEEEtran}
\bibliography{IEEEabrv, References}

\end{document}